\documentclass[11pt, a4paper]{article}

\usepackage[a4paper, top=2.5cm, bottom=2.5cm, left=2cm, right=2cm]{geometry}
\usepackage{fontspec}
\usepackage[english]{babel}

\usepackage{amsmath}
\usepackage{amssymb}
\usepackage{xcolor}
\usepackage{hyperref}
\usepackage{graphicx}
\usepackage{subcaption}
\usepackage{booktabs}
\usepackage{cancel}
\usepackage{authblk}

\newcommand{\R}{\mathbb{R}}
\newcommand{\norm}[1]{\left\lVert#1\right\rVert}
\newcommand{\E}{\mathbb{E}}
\newcommand{\Var}{\mathrm{Var}}
\newcommand{\Cov}{\mathrm{Cov}}
\newcommand{\Tr}{\mathrm{Tr}}
\newcommand{\Snoise}{\Sigma_{\mathrm{noise}}}
\newcommand{\tSnoise}{\widetilde{\Sigma}_{\mathrm{noise}}}
\newcommand{\Sinf}{\Sigma_\infty}
\newcommand{\Ssgd}{\Sigma_\infty^{\mathrm{(SGD)}}}
\newcommand{\Sreset}{\Sigma_\infty^{\mathrm{(reset)}}}
\newcommand{\minf}{m_\infty}
\newcommand{\Minf}{M_\infty}
\newcommand{\wridge}{w_{\mathrm{ridge}}}

\title{Ridge Regression from Poisson Resetting: A Renewal Perspective on Spectral Regularization}

\author{Petar Jolakoski\thanks{\texttt{jolakoskip@manu.edu.mk; petar.jolak@gmail.com}}}
\affil{\footnotesize Research Center for Computer Science and Information Technologies, Macedonian Academy of Sciences and Arts, Bul. Krste Misirkov 2, 1000 Skopje, Macedonia}

\date{May 28, 2026}

\begin{document}

\maketitle

\begin{abstract}
We connect stochastic resetting from non-equilibrium statistical physics with ridge regularization in statistical learning. For linear gradient flow, resetting to the origin at rate $r$ produces stationary mean $(X^\top X+rI)^{-1}X^\top y$, exactly the ridge estimator with penalty $\lambda=r$. This uses the known Laplace-transform relationship between ridge regression and exponential-time averaging of gradient flow, with the exponential time now interpreted as the stationary age associated with Poisson resetting. We then extend this identity to general renewal reset laws: the exponential reset time distribution is the unique renewal law whose stationary mean reproduces scalar ridge in every eigendirection as an exact filter identity for every positive curvature, while non-exponential renewal laws generate alternative spectral filters. At the fluctuation level, we study a separate additive Ornstein-Uhlenbeck extension with constant diffusion, interpreted as a stylized SGD approximation. In this setting, the equality holds only at the level of the mean, since the reset process has a nonzero stationary covariance from accumulated OU noise and reset-timing variance, whereas deterministic ridge is a fixed estimator with the same center. Stylized experiments compare the deterministic renewal-induced filters directly and illustrate when filters induced by non-exponential reset-time laws can differ predictively from ridge. The results for the stationary mean and the induced spectral filters are established for continuous-time gradient flow with isotropic resetting on quadratic objectives; the covariance and risk formulas additionally assume additive noise with state-independent covariance.
\end{abstract}

\section{Introduction}

In statistical learning, regularization controls the fundamental trade-off between model complexity and generalization. In least-squares estimation, highly correlated predictors or weakly identified directions can make the fitted coefficients highly sensitive to small perturbations in the training data. This instability can lead to unreliable out-of-sample predictions when new observations depend on directions that were only weakly constrained during training. Ridge regression addresses this instability by applying a simple quadratic regularization to the least-squares estimator. Specifically, it replaces the unstable inverse of \(X^\top X\) by the shifted normal matrix \(X^\top X+\lambda I\) and its inverse resolvent \((X^\top X+\lambda I)^{-1}\), shrinking each eigendirection by \(\mu/(\mu+\lambda)\) and damping weakly identified modes while largely preserving well-identified ones~\cite{hoerl1970ridge}. Subsequent sampling-theoretic work showed that this bias can improve mean-squared error under broader quadratic criteria~\cite{theobald1974generalizations}, and that analogous MSE comparisons still apply when \(X^\top X\) is singular, by comparing ridge with the generalized-inverse estimator for identifiable linear combinations of the coefficients~\cite{farebrother1976further}. For neural networks, the same explicit \(\ell_2\) mechanism appears as weight decay~\cite{krogh1991simple}. More broadly, inverse-problem and learning-theory work treats ridge as one point inside a larger class of spectral methods whose behaviour is determined by filter shape rather than by a single scalar penalty; Tikhonov regularization, Landweber iteration, spectral cut-off, and related procedures can all be cast in this common spectral-filter framework~\cite{rosasco2005spectral,bauer2007regularization,logerfo2008spectral}. A natural question here is which stochastic optimization dynamics reproduce the ridge filter exactly, and which instead generate different regularizers.

The standard comparison object is early stopping. In least-squares and kernel settings, gradient flow and Landweber iteration act as iterative regularizers whose stopping time induces exponential-type spectral filters~\cite{yao2007early,ali2019continuous}, while broader inverse-problem work places early stopping within general spectral-regularization and adaptive-stopping theory~\cite{blanchard2018optimal}. Recent work develops this distinction further by showing that early-stopped gradient-flow or gradient-descent solutions can be viewed as generalized ridge procedures with mode-dependent penalties~\cite{sonthalia2024early}, and that performance comparisons between tuned estimators depend strongly on filter class and spectral regime~\cite{dicker2017kernel,richards2021comparing,lu2024saturation}. More broadly, continuous-time analyses of accelerated methods and adaptive restart rules explain momentum-induced oscillations and show how deterministic restart can improve optimization without prior curvature knowledge, but these results concern momentum control rather than exact estimator identities of ridge type~\cite{su2016differential,odonoghue2015adaptive,giselsson2014monotonicity}. On the stochastic-optimization side, OU approximations of SGD characterize stationary covariances near quadratic minima~\cite{mandt2017stochastic}, while matched SDE approximations in high-dimensional quadratic problems characterize exact or asymptotic risk trajectories~\cite{paquette2022implicit}. For least squares, stochastic gradient flow analyses come closer to ridge at the mean or bias level, but still pay an additional variance price from mini-batching~\cite{ali2020implicit}. Other SGD analyses instead emphasize a distinct implicit spectral profile that can be competitive with or improve on ridge in favorable regimes~\cite{zou2021benefits}. Other implicit-regularization results identify gradient-, Jacobian-, or data-dependent anisotropic biases~\cite{barrett2020implicit,blanc2020implicit}. 

A complementary area of research in non-equilibrium statistical physics studies stochastic resetting: a process evolves according to its native dynamics until a reset event occurs, at which point it is returned to a reference state. Since Evans and Majumdar~\cite{evans2011diffusion}, Poisson resetting has served as a canonical mechanism for generating nonequilibrium stationary states and optimizing first-passage times~\cite{evans2020stochastic}. Beyond Poisson resetting, restart and renewal theory studies how different reset-time distributions change the behavior of reset processes. Arbitrary restart distributions admit universal renewal formulas, deterministic restart can outperform stochastic restart for mean completion time, and constructive performance bounds can be derived beyond the Poisson case~\cite{pal2017first,starkov2023universal,nikitin2024constructing}. Random search with resetting likewise emphasizes the central role of renewal structure~\cite{chechkin2018random}, and recent work has extended the framework to adaptive, state-dependent resetting and data-driven design of reset protocols~\cite{keidar2025adaptive}. Machine learning literature contains related restart heuristics, including learning-rate warm restarts in SGDR~\cite{loshchilov2017sgdr}, periodic partial parameter resets to mitigate primacy bias in deep reinforcement learning~\cite{nikishin2022primacy}, first-passage based optimization of training perturbations~\cite{meir2025first}, stochastic checkpoint resets under label noise~\cite{bae2025stochastic}, and soft OU-style resets toward initialization in non-stationary learning~\cite{galashov2024non}.

The link between ridge and randomized time averaging was noted by Tibshirani~\cite{tibshirani2022laplace}, who showed that the ridge solution can be written as a Laplace transform of the gradient-flow path, or equivalently as the expectation of the gradient-flow estimator at an exponential random time. We interpret this exponential time as the equilibrium age associated with Poisson resetting, whose reset intervals are exponentially distributed. Under that interpretation, ridge appears as the stationary mean of reset gradient flow, and changing the reset-time law changes the spectral filter through the corresponding equilibrium-age distribution. For general reset-interval laws, the same calculation yields a law-dependent spectral filter. It also separates the induced mean estimator from the extra fluctuations introduced by the reset dynamics.

The paper is organized as follows. Sections~\ref{sec:linear} and~\ref{sec:laplace_identity} establish the exact Poisson-ridge identity and its Laplace-averaging mechanism. Section~\ref{sec:renewal_main} extends this calculation to general renewal reset laws and characterizes their induced filters, including the uniqueness of exponential resets when equality with ridge is required for every \(\mu>0\). Sections~\ref{sec:covariance}-\ref{sec:renewal_risk_beyond_mean} derive the variance introduced when a reset process is used to realize one of these mean filters. The empirical section applies the induced filters as deterministic spectral estimators, so that the comparisons concern filter shape rather than trajectory fluctuations.

\section{Linear Models: Exact Equivalence in the Stationary Mean}
\label{sec:linear}

Consider the standard linear regression objective
\begin{equation}
J(w) = \frac12 \norm{Xw-y}^2,
\label{eq:loss_linear_ped}
\end{equation}
where \(X \in \R^{n \times d}\) is the design matrix, \(y \in \R^n\), and \(w \in \R^d\). Define\footnote{Throughout the paper, uppercase letters such as \(H\), \(X\), \(V\), and \(\Sigma\) usually denote matrices, lowercase Roman letters such as \(w\), \(m\), and \(b\) usually denote vectors, and scalar symbols such as \(r\), \(\mu\), and \(\lambda\) usually denote scalar quantities. The convention is only meant to aid readability; locally defined quantities retain their stated meanings.}
\begin{equation}
H:=X^\top X, \qquad b:=X^\top y.
\label{eq:H_b_defs_ped}
\end{equation}
Then \(\nabla J(w)=Hw-b\), and because \(J\) is quadratic, \(H\) is constant and positive semidefinite. The next step is to express ordinary training in continuous time. Gradient flow, the continuous-time limit of gradient descent, obeys the \(d\)-dimensional linear ODE system
\begin{equation}
\frac{dw}{dt} = -\nabla J(w)= -Hw + b.
\label{eq:gf_linear_ped}
\end{equation}
Now augment gradient flow with Poisson resetting to \(0\) at rate \(r>0\). To derive the corresponding mean equation, consider a small interval \([t,t+\Delta t]\). Over this interval, either no reset occurs, with probability \(1-r\Delta t+o(\Delta t)\), and the state evolves according to the drift \(-Hw+b\), or one reset occurs, with probability \(r\Delta t+o(\Delta t)\), and the state is returned to \(0\). Writing \(m(t):=\E[w(t)]\), this gives
\[
m(t+\Delta t)
=(1-r\Delta t)\bigl(m(t)+(-Hm(t)+b)\Delta t\bigr)
+r\Delta t\cdot 0
+o(\Delta t).
\]
Subtracting \(m(t)\), dividing by \(\Delta t\), and letting \(\Delta t\to 0\) yields a \(d\)-dimensional linear ODE system for the mean vector,
\begin{equation}
\frac{dm}{dt}=-(H+rI)m+b.
\label{eq:mean_dynamics_ped}
\end{equation}
The interpretation is simple: \(-Hm+b\) is the usual gradient-flow drift, and the extra term \(-rm\) is the average pull toward the reset point \(0\). A step-by-step version of this conditioning argument is given in Appendix~\ref{app:pedagogical_mean}. Since \(H\succeq 0\) and \(r>0\), the shifted matrix \(H+rI\) is positive definite and therefore invertible. Solving the matrix ODE \eqref{eq:mean_dynamics_ped} via the matrix exponential gives
\begin{equation}
m(t)=e^{-(H+rI)t}m(0) + (H+rI)^{-1}\big(I-e^{-(H+rI)t}\big)b.
\label{eq:mean_transient_solution_ped}
\end{equation}
\noindent and therefore
\begin{equation}
m_\infty := \lim_{t\to\infty} m(t) = (H+rI)^{-1}b.
\label{eq:reset_stationary_ped_pre}
\end{equation}
Substituting \(H=X^\top X\) and \(b=X^\top y\),
\begin{equation}
m_\infty = (X^\top X + rI)^{-1}X^\top y.
\label{eq:reset_stationary_ped}
\end{equation}
Recall that ridge regression with penalty \(\lambda\) has the well-known closed-form solution~\cite{hoerl1970ridge} (see Appendix~\ref{app:pedagogical_ridge} for a self-contained derivation)
\begin{equation}
w_{\text{ridge}} = (X^\top X + \lambda I)^{-1}X^\top y.
\label{eq:ridge_solution_ped}
\end{equation}
The ridge identification is now immediate from \eqref{eq:reset_stationary_ped}. Comparing
\[
m_\infty=(X^\top X+rI)^{-1}X^\top y
\]
with the ridge formula \eqref{eq:ridge_solution_ped} gives
\[
m_\infty \equiv w_{\mathrm{ridge}}
\qquad \text{when } \lambda=r.
\]
Thus, in linear quadratic models, Poisson resetting inserts exactly the same isotropic shift \(rI\) into the normal equations that ridge uses.

\begin{figure}[htbp]
\centering
\includegraphics[width=11cm]{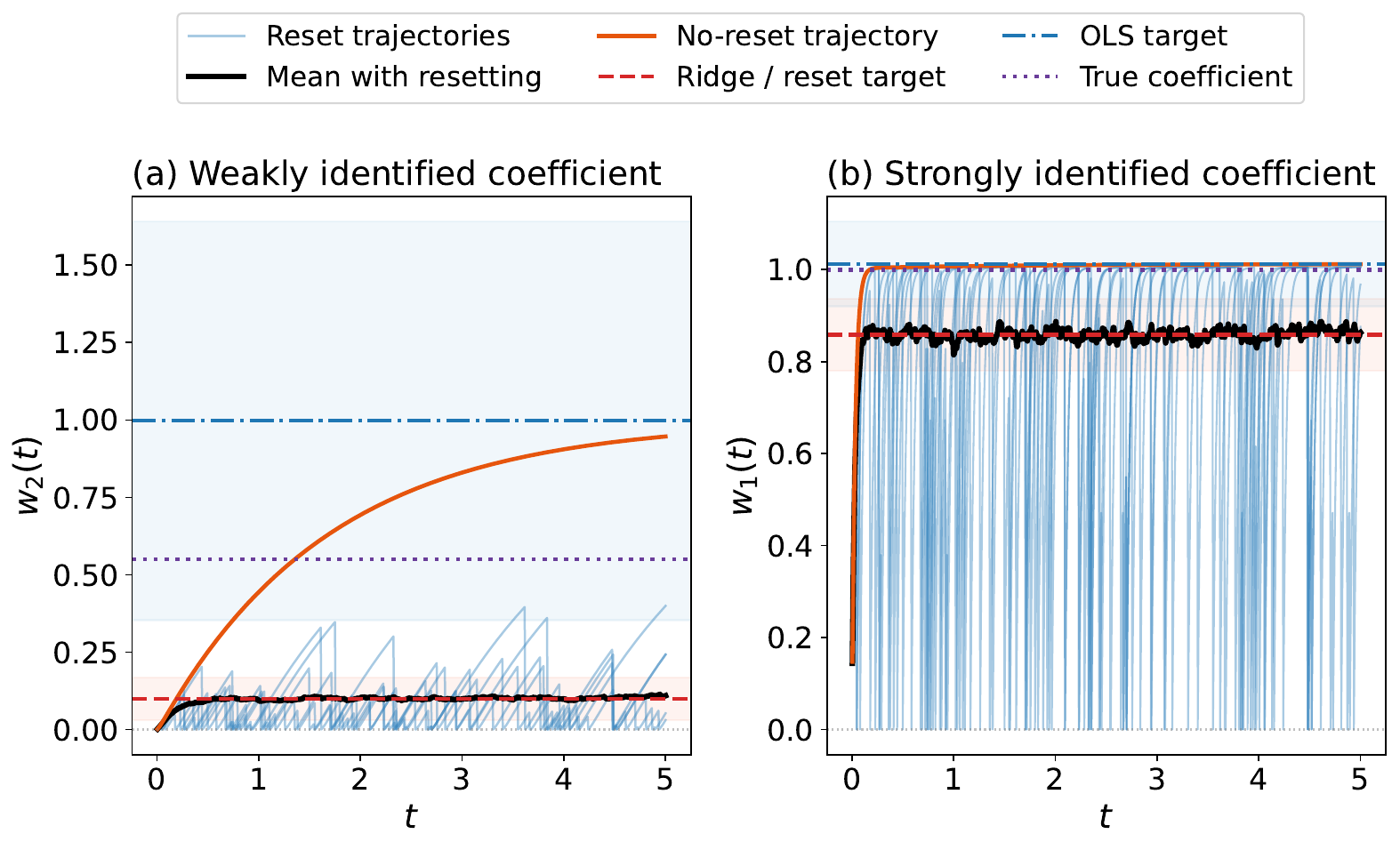}
\caption{\textbf{Gradient-flow trajectories with and without Poisson resetting in a two-parameter linear regression problem.} The left panel shows a weakly identified coefficient, corresponding to a predictor with low variance, while the right panel shows a strongly identified coefficient. Blue curves are trajectories with resetting at rate $r$, the black curve is their ensemble mean, and the orange curve is the corresponding trajectory without resetting. In each panel, the dashed red line is the ridge/reset target $(X^\top X + rI)^{-1}X^\top y$, the dash-dotted blue line is the minimum-norm OLS solution $(X^\top X)^{+}X^\top y$, and the dotted purple line is the true coefficient used to generate the data. The shaded red and blue bands indicate $\pm 2$ standard errors for the ridge and OLS estimators, respectively, computed under the classical model for observation noise conditional on the displayed design \(X\); they are not reset-path uncertainty bands. Resetting tracks the ridge solution and has its largest visible effect on the weakly identified coefficient, where OLS is most variable.}
\label{fig:mean_equivalence}
\end{figure}

Figure~\ref{fig:mean_equivalence} shows the distinction between sample paths and the averaged estimator. Individual trajectories continue to jump and restart, and no single path equals the ridge solution. The exact match appears only after averaging over the reset time distribution. Thus, the identity has a specific scope such that Poisson resetting reproduces ridge only at the level of the stationary mean.

Two short remarks are useful here. First, resetting to a nonzero reference point \(w_0\) gives \(m_\infty=(H+rI)^{-1}(b+rw_0)\), namely ridge centered at \(w_0\). Second, under the normalized loss \(\frac{1}{2n}\|Xw-y\|^2\), the same calculation gives \(m_\infty=(X^\top X+nrI)^{-1}X^\top y\). If ridge is written with the same normalized data-fit term, \(\frac{1}{2n}\|Xw-y\|^2+\frac{\lambda}{2}\|w\|^2\), then the matching parameter is still \(\lambda=r\); equivalently, relative to the unnormalized convention in \eqref{eq:ridge_solution_ped}, the same estimator can be written with penalty \(nr\).

\subsection{Spectral Interpretation}
\label{sec:spectral_interpretation_linear}

Let \(H = V\Lambda V^\top\) with \(\Lambda=\mathrm{diag}(\mu_1,\dots,\mu_d)\) and define spectral coordinates \(\tilde m_\infty := V^\top m_\infty\), \(\tilde b := V^\top b\), and \(\tilde w^\star := V^\top w^\star\), where \(w^\star:=H^+b\) is the minimum-norm OLS solution. Because \(b=X^\top y\) lies in the column space of $H$, every zero-eigenvalue mode also satisfies \(\tilde b_i=0\). Working in the eigenbasis (see Appendix~\ref{app:pedagogical_spectral} for the detailed derivation), each mode of the stationary mean satisfies
\begin{equation}
\tilde{m}_{\infty, i} = \frac{\mu_i}{\mu_i + r} \tilde{w}^\star_i,
\label{eq:shrinkage}
\end{equation}
where $\tilde{w}^\star_i$ are the coordinates of the unregularized minimum-norm OLS solution.

As a result, the stationary mean has the usual ridge-shrinkage form. When \(\mu_i \gg r\), relaxation in mode \(i\) is fast relative to the reset timescale and the mode is largely preserved. When \(\mu_i \ll r\), resets interrupt relaxation repeatedly and the mode is strongly suppressed. In other words, resetting is a timescale competition between gradient flow and renewal.

Figure~\ref{fig:mean_equivalence} makes the spectral interpretation intuitive. The left panel corresponds to a weakly identified direction where the dataset does not determine that coefficient very precisely, so many nearby values fit almost equally well and the OLS estimate in that direction is unstable. In practice, this is exactly the regime where noise, collinearity, or low feature variation can move the unregularized estimate substantially. Because progress toward that unstable OLS target is slow between reset events, the average effect of resetting is strong, and the gap between the ridge/reset level and the OLS level is visibly large. On the other hand, the right panel corresponds to a well-identified direction where the data are much more informative about that coefficient, so its unregularized estimate is more stable and the trajectory gets close to it quickly between resets. As a result, resetting changes that coefficient much less. In other words, resetting acts most strongly on coefficients that the data do not determine reliably, while leaving well-supported coefficients much closer to their unregularized values.

\section{Ridge as an Exponential-Time Average of Gradient Flow}
\label{sec:laplace_identity}

The previous section established the mean identity by solving the linear ODE system for the mean vector. Here, we show a derivation using the equilibrium-age distribution. In particular, under Poisson resetting, the stationary mean is a Laplace average of the no-reset gradient-flow trajectory. This is the same renewal mechanism by which Poisson resetting produces stationary states in the statistical physics literature~\cite{evans2011diffusion,evans2020stochastic}.

Recall the gradient-flow ODE from Section~\ref{sec:linear},
\begin{equation}
\dot \beta(t) = -H\beta(t) + b,
\qquad \beta(0)=0,
\label{eq:gf_zero_init_laplace}
\end{equation}
If \(A\) denotes the stationary age since the last Poisson reset, then \(A\sim \mathrm{Exp}(r)\) by memorylessness:
\begin{equation}
A \sim \mathrm{Exp}(r),
\qquad
f_A(a) = r\,e^{-ra}, \quad a \ge 0.
\label{eq:stationary_age_exp}
\end{equation}
At stationarity, the current state is obtained by following the no-reset gradient-flow path from \(0\) for the elapsed time \(A\) since the most recent reset. Conditioning on \(A\) gives the state \(\beta(A)\), so the stationary mean is
\begin{equation}
m_\infty(r)
= \E[\beta(A)]
= r\int_0^\infty e^{-ra}\,\beta(a)\,da.
\label{eq:laplace_average_identity_main}
\end{equation}
Poisson resetting is not equivalent to stopping gradient flow at one fixed time. Instead, it averages the no-reset trajectory over all elapsed times with exponential weights.

Variation of constants gives
\begin{equation}
\beta(t)=\int_0^t e^{-H(t-s)}\,b\,ds.
\label{eq:beta_integral_rep_laplace}
\end{equation}
\noindent Substituting \eqref{eq:beta_integral_rep_laplace} into \eqref{eq:laplace_average_identity_main} gives
\begin{equation}
m_\infty(r)
= r\int_0^\infty e^{-rt}\left(\int_0^t e^{-H(t-s)}\,b\,ds\right)dt.
\label{eq:m_inf_double_integral_start}
\end{equation}
\noindent Because \(r>0\), \(H\succeq 0\), and \(\|e^{-H(t-s)}b\|\le \|b\|\), the double integral is absolutely integrable, so Fubini's theorem applies. Exchanging the order of integration and setting \(u=t-s\) yields
\begin{align}
m_\infty(r)
&= r\int_0^\infty \left(\int_{s}^{\infty} e^{-rt}\,e^{-H(t-s)}\,dt\right)b\,ds \label{eq:m_inf_swap_order}\\
&= r\int_0^\infty e^{-rs}\,ds \cdot \int_0^\infty e^{-(H+rI)u}\,du\,\cdot\, b, \label{eq:m_inf_separate_integrals}
\end{align}
\noindent so
\begin{equation}
m_\infty(r)=\int_0^\infty e^{-(H+rI)u}\,du\;\cdot\; b=(H+rI)^{-1}b.
\label{eq:laplace_averaging_theorem_statement}
\end{equation}
This is the mechanism behind the reset-ridge correspondence. The exponential age density turns the matrix exponential \(e^{-tH}\) into the shifted resolvent, reproducing the known Laplace-transform identity between gradient flow and ridge regression~\cite{tibshirani2022laplace}. The previous section described the effect mode by mode. The present section derives the same resolvent directly from the renewal-time average.

Substituting \(H=X^\top X\) and \(b=X^\top y\), the same calculation gives
\begin{equation}
m_\infty(r)=(X^\top X+rI)^{-1}X^\top y = w_{\mathrm{ridge}}(\lambda)\quad \text{when } \lambda=r,
\label{eq:ridge_reset_corollary_laplace}
\end{equation}
recovering the stationary-mean identity from a more structural point of view. The derivation also shows why Poisson resetting is special: its reset intervals are exponentially distributed, and changing the reset time distribution changes the equilibrium-age law and hence the spectral filter.

\section{Spectral Filters from Renewal Resetting}
\label{sec:renewal_main}

The Poisson case extends naturally once reset intervals are allowed to follow a different renewal law. In that setting, the stationary mean is obtained by averaging over the equilibrium age since the most recent reset, and the Laplace transform of that age determines the associated spectral shrinkage filter. Exponential intervals reproduce ridge, whereas non-exponential intervals lead to different shrinkage profiles. Renewal formulations of this kind are standard in the resetting literature~\cite{evans2020stochastic,chechkin2018random}.

For a general renewal law, the formulas below describe the process as seen at a generic large observation time, chosen independently of the reset schedule. Thus \(A\) is the elapsed time since the most recent reset at that observation time. For non-arithmetic renewal laws, this gives the usual limiting age distribution. For deterministic resetting \(S\equiv T\), it corresponds to averaging uniformly over the phase within one reset cycle.

Let \(S\) denote a reset time distribution with \(\E[S]<\infty\) and \(S>0\) almost surely. Under the generic large-time observation convention just described, let \(A\) denote the corresponding equilibrium age, namely the elapsed time since the last reset. Between resets the state follows the same deterministic gradient-flow trajectory
\[
\phi(a)=\int_0^a e^{-Hu}b\,du
\]
used in Section~\ref{sec:laplace_identity}. If \(w^\star:=H^+b\) denotes the minimum-norm OLS solution, then in the eigenbasis of \(H\) this trajectory satisfies \(\tilde\phi_i(a)=(1-e^{-\mu_i a})\tilde w_i^\star\). Applying the renewal-reward theorem (see, for example, Ross~\cite{ross2014introduction}) to the indicator reward \(\mathbf{1}\{A(t)\le a\}\) gives the standard equilibrium-age distribution~\cite{gallager1996discrete}
\begin{equation}
\Pr(A \le a)=\frac{1}{\E[S]}\int_0^a \Pr(S>u)\,du,
\label{eq:equilibrium_age_cdf_main}
\end{equation}
because the reward accumulated in one reset interval is \(\min(S,a)\), whose expectation equals the displayed tail integral. Consequently, for every \(\mu>0\),
\begin{equation}
\E[e^{-\mu A}]
=
\frac{1-\mathcal L_S(\mu)}{\mu\,\E[S]},
\qquad
\mathcal L_S(\mu):=\E[e^{-\mu S}],
\label{eq:equilibrium_age_laplace_main}
\end{equation}
This identity follows from the equilibrium-age formula by integration by parts. Conditioning on \(A\) gives the stationary mean directly:
\[
m_\infty^{(S)}
=
\E[\phi(A)]
\]
\noindent In the eigenbasis of \(H\), the stationary mean has the modewise form
\[
\tilde m_{\infty,i}^{(S)}
=
\bigl(1-\E[e^{-\mu_i A}]\bigr)\tilde w_i^\star
=
g_S(\mu_i)\,\tilde w_i^\star,
\qquad
g_S(\mu)=1-\frac{1-\mathcal L_S(\mu)}{\mu\,\E[S]}.
\]
At zero curvature, we define \(g_S(0)=0\). This matches the fact that a zero-curvature direction has no gradient-flow movement from the zero initialization; equivalently, since \(b\) lies in the column space of \(H\), both the projected right-hand side \(\tilde b_i\) and the minimum-norm solution coordinate \(\tilde w_i^\star\) vanish in every nullspace direction of \(H\). With this convention, renewal resetting replaces a fixed stopping time by a random equilibrium age and induces a spectral filter through that age law. The resulting mean estimator depends on the full age distribution, so two reset laws with the same average reset interval can still produce different filters.

The same formula isolates why the exponential reset time distribution is unique. If one insists that \(g_S(\mu)=\mu/(\mu+\lambda)\) for every \(\mu>0\), then
\[
\frac{1-\mathcal L_S(\mu)}{\mu\,\E[S]}=\frac{\lambda}{\mu+\lambda},
\qquad
1-\mathcal L_S(\mu)=\E[S]\lambda\,\frac{\mu}{\mu+\lambda}.
\]
\noindent Since \(S>0\) almost surely, \(\mathcal L_S(\mu)\to 0\) as \(\mu\to\infty\), necessarily \(\E[S]\lambda=1\). Substituting back yields
\[
\mathcal L_S(\mu)=\frac{\lambda}{\mu+\lambda},
\]
which is exactly the Laplace transform of an exponential law. Within the renewal family, only exponentially distributed reset intervals reproduce scalar ridge in every eigendirection. This uniqueness statement concerns the whole filter function. It says that if a renewal law matches the scalar ridge filter \(g(\mu)=\mu/(\mu+\lambda)\) for every \(\mu>0\), then the reset intervals must be exponential. For one fixed finite design, however, the filter is only evaluated at the finitely many eigenvalues of \(H\). A non-exponential law could in principle agree with ridge at those finitely many points while differing at other curvatures.

\subsection{Deterministic periodic resetting}

The simplest non-exponential example is deterministic periodic resetting. In this case resets occur exactly every \(T\) units of time, rather than after random waiting times. Equivalently, the reset-time law is \(S\equiv T\). The age at a generic observation time is then uniform on \([0,T]\), so the cycle-averaged mean satisfies (see App.~\ref{app:renewal_mean_periodic} for more details):
\begin{equation}
g_T(\mu)=1-\frac{1-e^{-\mu T}}{\mu T},
\qquad
\tilde{\bar m}_{T,i} = g_T(\mu_i)\,\tilde w_i^\star.
\label{eq:periodic_filter_main}
\end{equation}
Because deterministic resetting is not memoryless, the continuous-time dynamics do not converge to a time-homogeneous stationary distribution; instead they approach a periodic steady state. Equation~\eqref{eq:periodic_filter_main} gives the interval-averaged mean, equivalently the mean obtained by observing the system at a uniformly random phase within a reset interval.

Expanding \eqref{eq:periodic_filter_main} for small \(\mu\) gives
\[
g_T(\mu)=\frac{\mu T}{2}-\frac{\mu^2T^2}{6}+O(\mu^3),
\]
\noindent whereas ridge gives
\[
\frac{\mu}{\mu+\lambda}
=
\frac{\mu}{\lambda}-\frac{\mu^2}{\lambda^2}+O(\mu^3).
\]
Matching the linear terms forces \(\lambda=2/T\), but the quadratic coefficients still disagree, so periodic resetting cannot be represented by any single ridge penalty across all modes. This mismatch shows that the reset-time law affects the full shape of the spectral filter. The next subsection introduces a one-parameter family that moves continuously from the exponential law to the periodic limit.

\subsection{Gamma/Erlang resetting}

To interpolate between exponential and periodic reset-time laws while keeping the mean reset interval fixed, take the reset interval \(S\) to have a Gamma distribution with shape \(k>0\) and scale \(\theta=\tau/k\). Then \(\E[S]=\tau\), while \(\mathrm{Var}(S)=\tau^2/k\), so increasing \(k\) makes the reset intervals more concentrated around their mean. The case \(k=1\) is the exponential law corresponding to Poisson resetting, and the limit \(k\to\infty\) collapses to deterministic resetting with period \(\tau\). The renewal filter depends on this reset-time law through the Laplace transform of \(S\), evaluated at the curvature \(\mu\). For the Gamma law, whose mean-filter derivation is given in App.~\ref{app:renewal_mean_gamma},
\[
\mathcal L_S(\mu)=\E[e^{-\mu S}]=(1+\theta\mu)^{-k}=\left(1+\frac{\tau\mu}{k}\right)^{-k},
\]
and substituting this expression into the general renewal formula gives
\begin{equation}
g_{k,\tau}(\mu)
=
1-\frac{1-\left(1+\frac{\tau\mu}{k}\right)^{-k}}{\tau\mu}.
\label{eq:gamma_filter_main}
\end{equation}
The parameter \(k\) controls the regularity of the reset schedule while preserving the mean reset time \(\tau\).

\begin{figure}[htbp]
\centering
\includegraphics[width=11cm]{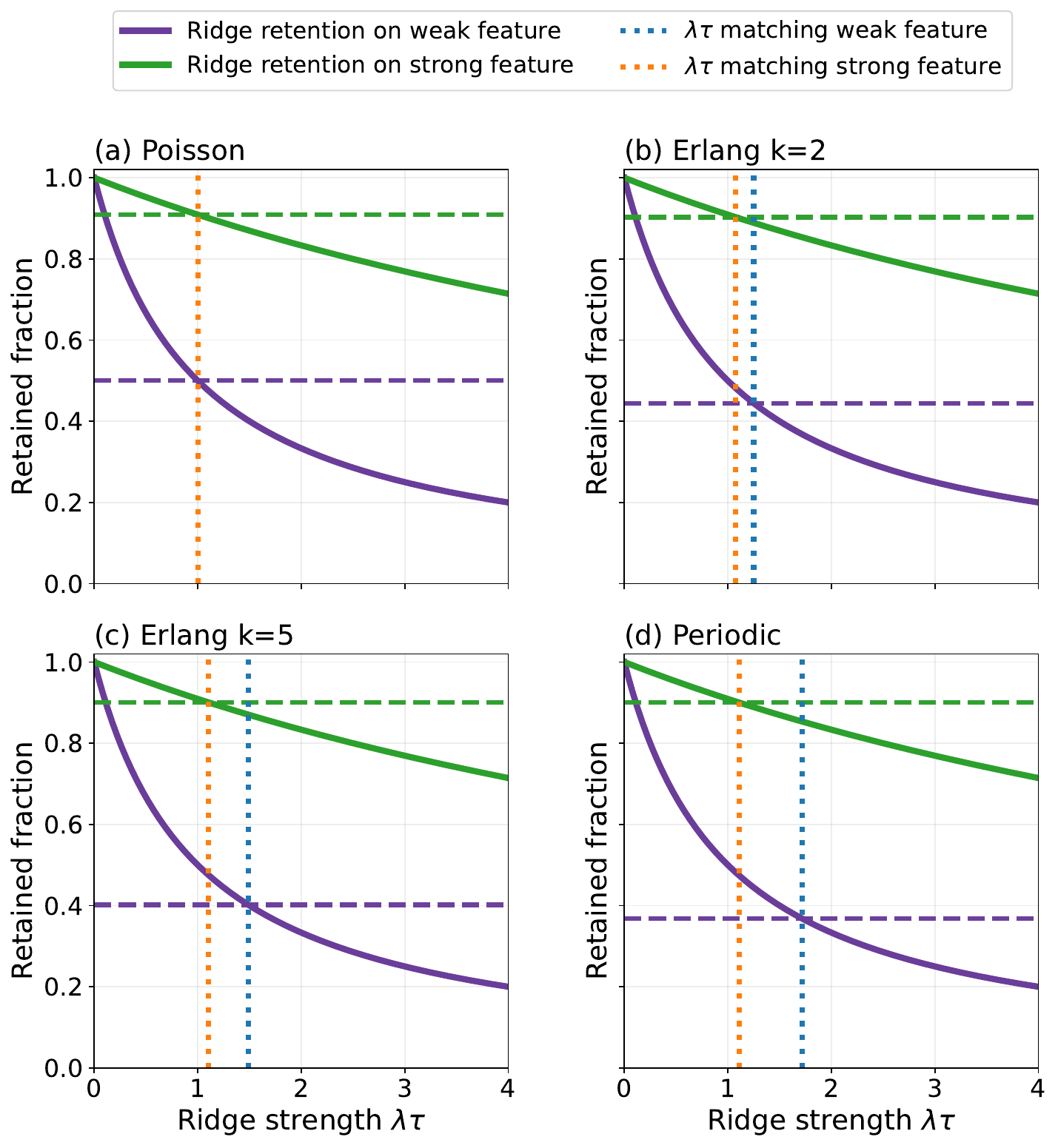}
\caption{\textbf{Two-mode comparison of ridge and renewal-induced shrinkage.} Each panel fixes two representative directions, a weak feature with \(\mu\tau=1\) and a strong feature with \(\mu\tau=10\), where \(\tau=\E[S]\) is the mean reset interval. The solid curves show the retained fraction that ordinary ridge would produce on those two features as the dimensionless ridge strength \(\lambda\tau\) varies. The dashed horizontal lines show the retained fractions generated by the indicated reset-time law, namely \(g_S(1)\) and \(g_S(10)\). The dotted vertical lines mark the ridge strengths needed to match the weak and strong features separately.}
\label{fig:renewal_ridge_mismatch}
\end{figure}

Figure~\ref{fig:renewal_ridge_mismatch} illustrates the effect of different renewal filters in a poorly conditioned linear quadratic problem. In particular, it compares how much of a weakly identified parameter and how much of a strongly identified parameter are kept after regularisation. The retained fraction is simply the learned coefficient divided by the corresponding no-regularisation coefficient, so in a mode with curvature \(\mu\) it is exactly the spectral filter value \(g(\mu)\). In each panel of Figure~\ref{fig:renewal_ridge_mismatch}, the purple and green curves show what ridge would retain on a weak and a strong feature as \(\lambda\tau\) is varied, while the horizontal dashed lines show what the chosen reset-time law retains at the same two values of \(\mu\tau\). The two vertical lines test whether the same value of \(\lambda\) matches the reset-induced shrinkage in both directions.

In the Poisson case, the exponential reset-time law induces exactly the ridge filter, so the two matching vertical lines coincide. The Erlang and periodic laws shown here behave differently, since no single ridge penalty matches the reset-induced shrinkage in both modes. Relative to ridge, these laws suppress the weak mode more strongly while retaining more of the strong mode. Whether that difference improves prediction depends on where signal and noise lie in the spectrum. For a fixed finite design, isolated coincidences at selected eigenvalues can occur, but the two-mode illustration reflects the stronger result above, namely that exact agreement with the ridge filter for every \(\mu>0\) is possible only under the exponential reset-time law.

The renewal viewpoint also constrains which spectral filters can be obtained by choosing a reset-time law. Writing \(h_S(\mu):=1-g_S(\mu)=\E[e^{-\mu A}]\), we see that the residual filter must be the Laplace transform of a probability law on \([0,\infty)\), hence completely monotone~\cite{schilling2012bernstein}. In particular, \(h_S(0)=1\), \(0\le h_S(\mu)\le1\), while the retained filter satisfies \(g_S(0)=0\) and \(g_S(\mu)\uparrow1\) as \(\mu\to\infty\). When \(\E[A]<\infty\),
\[
g_S(\mu)=\mu\E[A]+o(\mu)\qquad(\mu\downarrow0).
\]
If the reset law has finite mean \(\tau=\E[S]\), then \(h_S(\mu)=(1-\mathcal L_S(\mu))/(\mu\tau)\), so for ordinary laws with \(\mathcal L_S(\mu)\to0\), \(h_S(\mu)\sim(\mu\tau)^{-1}\) and \(g_S(\mu)=1-(\mu\tau)^{-1}+o(\mu^{-1})\) at high curvature. A ridge-equivalent effective penalty can be defined modewise by
\[
\lambda_{\mathrm{eff}}(\mu):=\mu\,\frac{h_S(\mu)}{1-h_S(\mu)},
\]
which is constant only for the exponential law. The \(1/\mu\) high-curvature residual tail gives the same first-order saturation order as ridge; non-exponential renewal laws change the mid-spectrum shape and low-curvature slope while keeping a smooth high-curvature tail. Conversely, a completely monotone residual \(h\) with \(h(0)=1\) arises from an equilibrium age of some renewal law with mean \(\tau\) precisely when \(1-\tau\mu h(\mu)\) is itself the Laplace transform of a probability law on \((0,\infty)\). Thus, renewal resetting generates a structured subclass of spectral filters, with ridge corresponding to the exponential member of that admissible family.

The next sections examine the extra variance that appears when the filter is realized dynamically rather than used as a deterministic estimator.

\section{Beyond the Mean: Variance Cost of Poisson Resetting}
\label{sec:covariance}

Sections\ref{sec:linear} and\ref{sec:laplace_identity} established that Poisson resetting reproduces ridge exactly at the level of the mean. At the process level, a single reset trajectory still fluctuates around the shared mean, so implementing ridge through random reset times carries an additional variance cost. To isolate that cost in a tractable setting, we now include stochastic gradient noise\footnote{Throughout the covariance and risk analysis, stochastic gradient noise means the additive OU approximation: conditional on \(X\), \(\Snoise\) is fixed and independent of \(w\), labels, mini-batch state, the Brownian motion, and the reset process. Exact finite-batch least-squares SGD generally has state-, label-, and mini-batch-dependent covariance.}, following standard diffusion approximations of SGD near quadratic minima, including OU-style local models and high-dimensional SDE risk descriptions~\cite{mandt2017stochastic,paquette2022implicit,ali2020implicit}, and, for tractability, model the dynamics as the \(d\)-dimensional Ornstein-Uhlenbeck (OU) process
\begin{equation}
dw(t) = (-Hw(t) + b)\,dt + \sigma\,dW(t),
\label{eq:ou_process}
\end{equation}
augmented with Poisson resetting to $w = 0$ at rate $r > 0$. Here $\sigma \in \R^{d \times d}$ is the diffusion matrix capturing stochastic gradient noise, $\Snoise := \sigma\sigma^\top$ is the noise covariance, and $W(t)$ is standard $d$-dimensional Brownian motion independent of the reset times.

To compute how diffusion noise and random resets affect the stationary mean and covariance, we use the infinitesimal generator of the reset process. For any test function \(\varphi\), the generator gives the instantaneous expected change of \(\varphi(w(t))\) when the current state is \(w\). For this piecewise-diffusion process with drift $F(w) = -Hw + b$, diffusion $\sigma$, and reset map $w \mapsto 0$ at rate $r$, the generator \(\mathcal{A}\) is defined by
\[
(\mathcal{A}\varphi)(w):=\lim_{h\downarrow0}\frac{\E[\varphi(w(h))\mid w(0)=w]-\varphi(w)}{h}.
\]
For a test function $\varphi \in C^2$, it takes the form
\begin{equation}
(\mathcal{A}\varphi)(w) = \nabla\varphi(w)^\top(-Hw + b) + \frac{1}{2}\Tr\!\bigl(\Snoise\,\nabla^2\varphi(w)\bigr) + r\bigl[\varphi(0) - \varphi(w)\bigr].
\label{eq:generator}
\end{equation}
At stationarity, $\E[\mathcal{A}\varphi(w)] = 0$ for all suitable test functions.

A key observation is that the diffusion noise does not affect the stationary mean. To see this, choose the test function $\varphi(w) = w_i$. Then $\nabla\varphi = e_i$, $\nabla^2\varphi = 0$, and $\varphi(0) = 0$, so
\[
(\mathcal{A}w_i)(w) = e_i^\top(-Hw + b) + \tfrac{1}{2}\Tr(\Snoise \cdot 0) + r(0 - w_i) = \bigl(-(H + rI)w + b\bigr)_i.
\]
The diffusion term vanishes identically because the Hessian of a linear function is zero. Setting $\E[\mathcal{A}w_i] = 0$ for all $i$ and using $H + rI \succ 0$ gives the unique solution
\begin{equation}
\minf = (H + rI)^{-1}b = \wridge(\lambda = r),
\label{eq:mean_noise_independent}
\end{equation}
confirming that the mean result from Section~\ref{sec:linear} holds unchanged in the presence of SGD noise.

\subsection{Stationary covariance and the modified Lyapunov equation}

Applying the generator to quadratic observables \(w_iw_j\) gives the \(d\times d\) matrix identity for the stationary second moment,
\begin{equation}
-H\Minf - \Minf H + b\minf^\top + \minf b^\top + \Snoise - r\Minf = 0,
\label{eq:second_moment_eq}
\end{equation}
with \(\Minf=\E[w_\infty w_\infty^\top]\in\R^{d\times d}\). Rewriting in terms of the stationary covariance \(\Sinf=\Minf-\minf\minf^\top\) yields
\begin{equation}
H\Sinf + \Sinf H + r\Sinf = r\minf\minf^\top + \Snoise.
\label{eq:cov_unsymmetrised}
\end{equation}
Equivalently,
\begin{equation}
\left(H + \frac{r}{2}\,I\right)\Sinf + \Sinf\left(H + \frac{r}{2}\,I\right) = r\,\minf\minf^\top + \Snoise.
\label{eq:lyapunov}
\end{equation}
This is a \(d\times d\) matrix Lyapunov equation, modified by Poisson resetting. Relative to the standard OU model, resetting adds a rank-one forcing term \(r\,\minf\minf^\top\), so the covariance is no longer independent of the mean.

Since \(H\) is symmetric, the equation decouples in its eigenbasis:
\[
\left(\mu_i + \frac{r}{2}\right)(\widetilde{\Sigma}_\infty)_{ij} + \left(\mu_j + \frac{r}{2}\right)(\widetilde{\Sigma}_\infty)_{ij} = r(\tilde{m}_\infty)_i(\tilde{m}_\infty)_j + (\tSnoise)_{ij},
\]
\noindent hence
\begin{equation}
(\widetilde{\Sigma}_\infty)_{ij} = \frac{r\,(\tilde{m}_\infty)_i\,(\tilde{m}_\infty)_j + (\tSnoise)_{ij}}{\mu_i + \mu_j + r}.
\label{eq:cov_eigenbasis}
\end{equation}
Appendix~\ref{app:renewal_verification} gives an independent renewal derivation of the same moment formulas.

\subsection{Covariance Decomposition}
\label{sec:decomposition}

\noindent Equation \eqref{eq:cov_eigenbasis} separates naturally into two pieces:
\begin{equation}
\Sinf = \Ssgd + \Sreset,
\label{eq:decomp}
\end{equation}
\noindent with eigenbasis entries
\begin{align}
(\widetilde{\Sigma}^{\mathrm{(SGD)}}_\infty)_{ij} &= \frac{(\tSnoise)_{ij}}{\mu_i + \mu_j + r}, \label{eq:sgd_component}\\[6pt]
(\widetilde{\Sigma}^{\mathrm{(reset)}}_\infty)_{ij} &= \frac{r\,\tilde{b}_i\,\tilde{b}_j}{(\mu_i + r)(\mu_j + r)(\mu_i + \mu_j + r)}. \label{eq:reset_component}
\end{align}
\(\Ssgd\) is accumulated diffusion noise. \(\Sreset\) is timing variance: it survives even when \(\sigma=0\), because different trajectories are observed at different deterministic positions between resets.

For the diagonal ($i = j$), the per-mode variance is:
\begin{equation}
\Var[\tilde{w}_{\infty,i}] = \underbrace{\frac{(\tSnoise)_{ii}}{2\mu_i + r}}_{\text{SGD piece}} + \underbrace{\frac{r\,\tilde{b}_i^2}{(\mu_i + r)^2(2\mu_i + r)}}_{\text{reset piece}}.
\label{eq:var_diagonal}
\end{equation}
In the original (non-eigenbasis) coordinates:
\[
\Ssgd = V\,\widetilde{\Sigma}^{\mathrm{(SGD)}}_\infty\,V^\top, \qquad \Sreset = V\,\widetilde{\Sigma}^{\mathrm{(reset)}}_\infty\,V^\top.
\]
Note that $\Ssgd$ and $\Sreset$ are generally not diagonal in the original basis, even though the eigenbasis entries have the clean componentwise form \eqref{eq:sgd_component}-\eqref{eq:reset_component}.

Both variance terms in \eqref{eq:var_diagonal} share the factor $1/(2\mu_i + r)$, so their ratio simplifies by cancellation:
\begin{equation}
\frac{\text{reset component}}{\text{SGD component}} = \frac{r\,\tilde{b}_i^2}{(\mu_i + r)^2\,(\tSnoise)_{ii}}.
\label{eq:snr_ratio}
\end{equation}
The relevant modewise signal-to-noise criterion is the full dimensionless ratio in \eqref{eq:snr_ratio}. The factor \(r/(\mu_i+r)^2\) accounts for the reset-relaxation timescale: it is large when reset events interrupt the mode before relaxation and small when relaxation is fast compared with resetting. As a result, modes with \(r\tilde b_i^2/[(\mu_i+r)^2(\tSnoise)_{ii}]\gg1\) are dominated by signal in this variance decomposition, so resets primarily erase useful deterministic progress. Modes with this ratio \(\ll1\) are dominated by diffusion, so resets mainly reduce the amount of OU noise that can accumulate between reset events. As a drift-to-diffusion ratio for each spectral mode, \eqref{eq:snr_ratio} plays the role of a modewise P\'eclet-type number.

\subsection{Risk decomposition and the variance cost of resetting}
\label{sec:risk}

This section addresses a different issue from the exact mean identity. Even when Poisson resetting reproduces ridge at the level of the mean, implementing that estimator through random reset times introduces additional fluctuations. To quantify the statistical cost of these fluctuations, we compare the resulting stationary estimator with its ridge counterpart under a standard fixed-design regression model.

Fix the design matrix \(X\) and write \(y=X\beta_0+\eta\) with \(\eta\sim(0,\sigma_\eta^2I)\). Averaging only over algorithmic randomness gives the conditional estimation risk
\begin{equation}
\mathcal{R}_{\mathrm{cond}}(r;\,X,y,\beta_0)
:=
\E_{\mathrm{alg}}\norm{w_\infty - \beta_0}^2
=
\norm{(H+rI)^{-1}b - \beta_0}^2 + \Tr(\Ssgd) + \Tr(\Sreset),
\label{eq:risk_decomp_cond}
\end{equation}
where
\begin{equation}
\Tr(\Ssgd) = \sum_{i=1}^d \frac{(\tSnoise)_{ii}}{2\mu_i + r},
\label{eq:tr_sgd}
\end{equation}
and
\begin{equation}
\Tr(\Sreset) = \sum_{i=1}^d \frac{r\,\tilde{b}_i^2}{(\mu_i + r)^2(2\mu_i + r)}.
\label{eq:tr_reset}
\end{equation}
The underlying identity is the usual bias-variance decomposition
\begin{equation}
\mathcal{R}_{\mathrm{cond}}(r;\,X,y,\beta_0)
=
\E_{\mathrm{alg}}\norm{w_\infty - \beta_0}^2
=
\norm{\minf - \beta_0}^2 + \Tr(\Sinf),
\label{eq:bias_var_decomp}
\end{equation}
\noindent together with \(\Sinf=\Ssgd+\Sreset\).

Averaging over the observation noise gives
\begin{equation}
\mathcal{R}(r;\,\beta_0)
:=
\E_\eta\!\left[\mathcal{R}_{\mathrm{cond}}(r;\,X,y,\beta_0)\right]
=
\mathrm{Risk}_{\mathrm{ridge}}(r;\,\beta_0) + \Tr(\Ssgd) + \E_\eta[\Tr(\Sreset)],
\label{eq:risk_decomp}
\end{equation}
where the first term is the ridge risk, consisting of squared bias and the variance of the ridge mean due to observation noise; Appendix~\ref{app:classical_ridge_risk} derives \eqref{eq:ridge_risk} from the classical ridge MSE formula of Hoerl and Kennard and the matrix MSE analyses of Theobald and Farebrother~\cite{hoerl1970ridge,theobald1974generalizations,farebrother1976further}:
\begin{equation}
\mathrm{Risk}_{\mathrm{ridge}}(r;\,\beta_0) = \sum_{i=1}^d \left[\frac{r^2\,(v_i^\top\beta_0)^2}{(\mu_i + r)^2} + \frac{\sigma_\eta^2\,\mu_i}{(\mu_i + r)^2}\right].
\label{eq:ridge_risk}
\end{equation}
The reset contribution after averaging over observation noise is
\begin{equation}
\E_\eta[\Tr(\Sreset)] = \sum_i \frac{r\,\E_\eta[\tilde{b}_i^2]}{(\mu_i+r)^2(2\mu_i+r)} = \sum_i \frac{r[\mu_i^2(v_i^\top\beta_0)^2 + \sigma_\eta^2\mu_i]}{(\mu_i+r)^2(2\mu_i+r)}.
\label{eq:reset_unconditional}
\end{equation}
Both risk notions are useful. The conditional risk describes one realised dataset, whereas \(\mathcal R(r;\beta_0)\) is the natural objective for choosing a reset rate within the Poisson family. Because the reset covariance formulas are derived for \(r>0\), and because boundary optima may occur only as \(r\downarrow0\) or \(r\to\infty\) in degenerate cases, we state the optimizer with an existence qualification. If the minimum is achieved at a finite positive reset rate, then
\begin{equation}
r^* \in \operatorname*{arg\,min}_{r > 0}\,\mathcal{R}(r;\,\beta_0).
\label{eq:total_risk}
\end{equation}
Otherwise the optimizer should be understood in the extended sense \(r^\star\in[0,\infty]\), with boundary cases \(r^\star=0\) or \(r^\star=\infty\) handled by the stated conventions.

\noindent For a fixed calibration \(r\), Poisson resetting reproduces the ridge mean, but adds two nonnegative fluctuation terms. Therefore
\begin{equation}
\mathcal{R}(r;\,\beta_0)\ge \mathrm{Risk}_{\mathrm{ridge}}(r;\,\beta_0),
\label{eq:variance_tax}
\end{equation}
with equality only in degenerate zero-variance limits. In that precise sense, exact ridge weakly dominates its Poisson reset implementation when the spectral filter is matched. At a matched filter, Poisson resetting does not improve risk over deterministic ridge.

\section{Beyond Poisson: Risk Across Renewal Laws}
\label{sec:renewal_risk_beyond_mean}

Once the objective shifts from exact mean matching to total risk, the matched comparison between Poisson resetting and ridge is already settled by \eqref{eq:variance_tax}. However, if one allows the reset-time law itself to change, how do the resulting filters induced by non-exponential reset-time laws trade mean shrinkage against accumulation of OU noise and reset timing variance?

\subsection{Exact renewal covariance and risk}

Let \(S\) denote the reset-time law and \(A\) the corresponding equilibrium age. Recall from Section~\ref{sec:renewal_main} that the mean filter is controlled by
\begin{equation}
h_S(\mu):=\E[e^{-\mu A}],
\qquad
g_S(\mu):=1-h_S(\mu).
\label{eq:main_renewal_h}
\end{equation}
Replacing the exponential equilibrium-age law of Poisson resetting by a general equilibrium-age law gives an exact covariance decomposition for the equilibrium snapshot distribution. In the eigenbasis of \(H\),
\begin{align}
\bigl(\widetilde{\Sigma}_S^{\mathrm{(SGD)}}\bigr)_{ij}
&=
(\tSnoise)_{ij}\,\frac{g_S(\mu_i+\mu_j)}{\mu_i+\mu_j},
\label{eq:main_renewal_sgd_component}\\[6pt]
\bigl(\widetilde{\Sigma}_S^{\mathrm{(tim)}}\bigr)_{ij}
&=
\tilde w_i^\star \tilde w_j^\star
\left[
h_S(\mu_i+\mu_j)-h_S(\mu_i)\,h_S(\mu_j)
\right].
\label{eq:main_renewal_timing_component}
\end{align}
The first term is accumulated OU noise over a random age; the second is pure timing variance from observing deterministic relaxation at a random point in the reset interval.

If \(y=X\beta_0+\eta\) with \(\eta\sim(0,\sigma_\eta^2I)\) and \(\alpha_i:=v_i^\top\beta_0\), the full snapshot risk after averaging over observation noise is
\begin{equation}
\mathrm{Risk}_{\mathrm{snap},S}(\beta_0)
=
\sum_{i=1}^d
\left[
h_S(\mu_i)^2\,\alpha_i^2
\;+\;
\frac{\sigma_\eta^2\,g_S(\mu_i)^2}{\mu_i}
\;+\;
\frac{(\tSnoise)_{ii}\,g_S(2\mu_i)}{2\mu_i}
\;+\;
\left(\alpha_i^2+\frac{\sigma_\eta^2}{\mu_i}\right)
\Bigl(h_S(2\mu_i)-h_S(\mu_i)^2\Bigr)
\right].
\label{eq:main_renewal_total_risk}
\end{equation}
Equation~\eqref{eq:main_renewal_total_risk} is the renewal law analogue of the Poisson three-term decomposition: risk of the mean estimator, accumulated SGD noise, and reset timing variance. Terms involving \(1/\mu_i\) are understood by continuous extension when \(\mu_i=0\), provided \(\E[A]<\infty\) (equivalently \(\E[S^2]<\infty\)). Then \(g_S(2\mu)/(2\mu)\to \E[A]\), while \(g_S(0)=0\) and \(h_S(0)=1\), so a nullspace mode contributes \(\alpha_i^2+(\tSnoise)_{ii}\E[A]\). In particular, it reduces to the irreducible bias term \(\alpha_i^2\) only when the algorithmic noise has no nullspace component. Appendix~\ref{app:renewal_cov_risk} derives these formulas in full and specializes them to periodic and Gamma/Erlang reset-time laws.

Changing the reset-time law changes three objects at once. First, \(g_S(\mu)\) sets the mean estimator and the tradeoff between bias and observation noise. Second, \(g_S(2\mu)/(2\mu)\) controls how much OU noise accumulates over a typical reset interval. Third, \(h_S(2\mu)-h_S(\mu)^2\) measures variance injected purely by uncertainty in the age at which the trajectory is observed.

At fixed mean reset interval \(\tau=\E[S]\), moving from the exponential law to sharper reset-time laws such as Erlang and periodic laws changes both shrinkage and fluctuation coefficients. The consequences depend on where each mode lies in the spectrum. For example, weak noisy modes can benefit because more regular reset-time laws suppress small-\(\mu\) directions more aggressively, reducing the contribution of poorly identified directions. On the other hand, strong signal-dominated modes can be harmed when this stronger shrinkage removes useful signal rather than mainly reducing noise.

\begin{figure}[htbp]
\centering
\includegraphics[width=\textwidth]{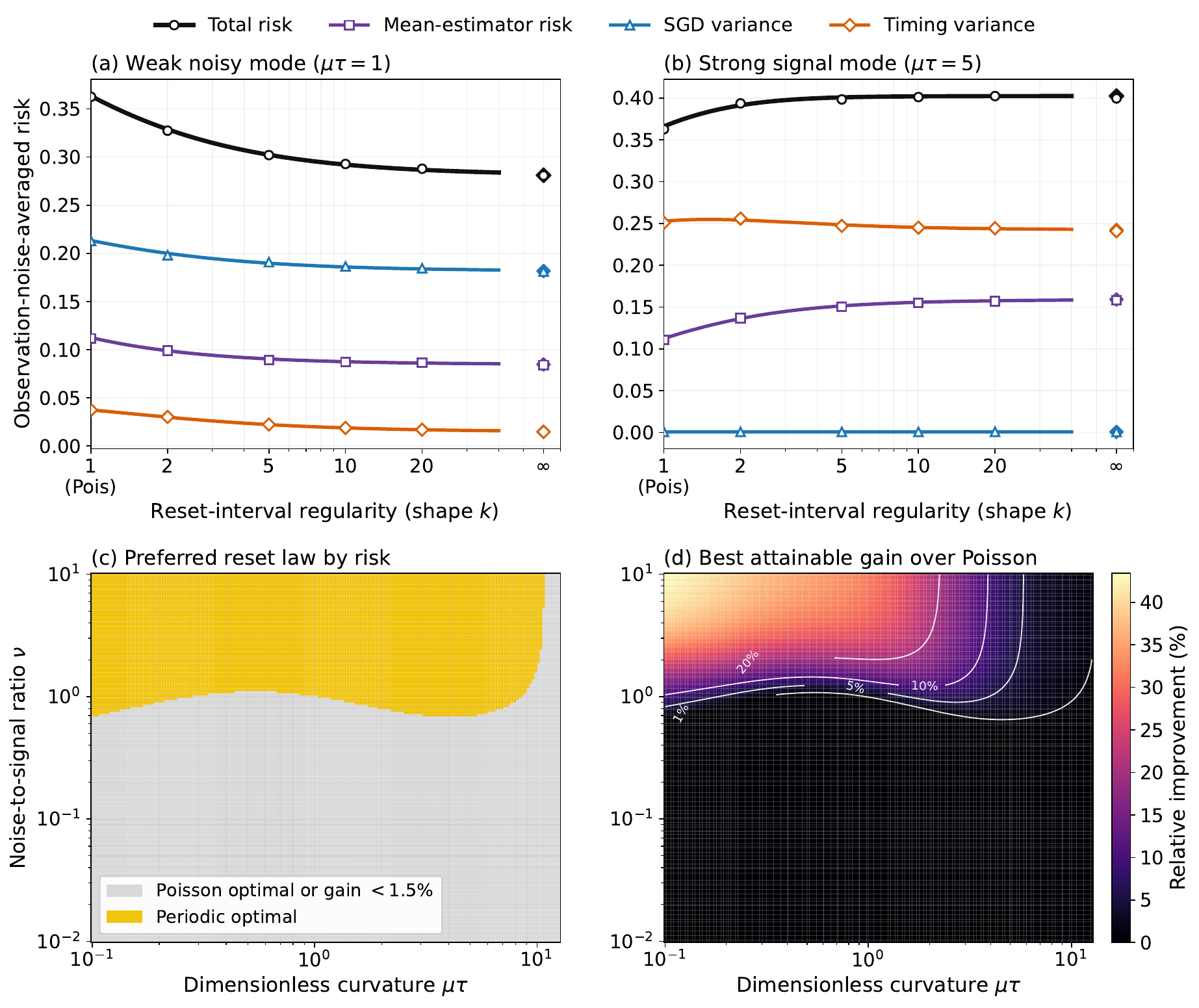}
\caption{\textbf{Renewal law risk beyond the mean.}
Top row: risk decomposition at fixed mean reset interval \(\tau\) as the reset-time law is varied from Poisson/exponential resetting (\(k=1\)) through Erlang laws to the periodic limit (\(k\to\infty\)). Solid curves show the exact renewal formulas after averaging over observation noise, and markers show empirical estimates from exact snapshots. Colors denote total risk (black), mean-estimator contribution (purple), accumulated SGD variance (blue), and timing variance (orange). The two panels compare a weak mode with \(\mu\tau=1\) and a strong mode with \(\mu\tau=5\). Bottom row: best reset-time law class and corresponding gain over Poisson resetting across curvature \(\mu\tau\) and effective joint noise level \(\nu\), with \(\beta_0=1\) and \(\sigma_\eta^2/\mu=\sigma_n^2\tau=\nu^2/2\). Gains below \(1.5\%\) are grouped with Poisson resetting.}
\label{fig:renewal_risk_landscape}
\end{figure}

Figure\ref{fig:renewal_risk_landscape} illustrates this tradeoff. Panel(a) shows a weak noisy mode with \(\mu\tau=1\), where increasing reset-time regularity lowers total risk monotonically from the Poisson-resetting value to the periodic limit. The contributions from the mean estimator, SGD, and timing all decrease, with the largest absolute drop coming from the SGD variance term. The simulation markers stay close to the theoretical curves, showing good agreement between theory and simulations. Panel~(b) shows the opposite regime. For the strong signal mode \(\mu\tau=5\), sharpening the reset-time law raises total risk because the increased risk of the mean estimator dominates the small reduction in timing variance, while the SGD contribution is negligible. Panel~(c) shows which reset-time law attains the lowest risk across dimensionless curvature \(\mu\tau\) and noise level. Improvements of less than \(1.5\%\) over Poisson resetting are treated as negligible and labelled as Poisson. Under this convention, Poisson resetting is preferred for low-noise, signal-dominated modes, whereas periodic resetting is preferred when noise is sufficiently large relative to signal. Finally, panel~(d) quantifies the size of that advantage. The best gain over Poisson resetting is concentrated in regimes with high noise and small to intermediate values of \(\mu\tau\), and disappears rapidly as modes have larger curvature or lower noise.

The covariance and risk formulas derived in Sections~\ref{sec:covariance}-\ref{sec:renewal_risk_beyond_mean} quantify the variance added by OU noise accumulation and by random age or phase sampling. The comparisons below do not simulate these reset trajectories. Instead, they apply the deterministic filter \(g_S(\mu)\) associated with each renewal law directly in the eigenbasis of the training Hessian. Therefore, their prediction errors compare the effect of filter shape alone and the fluctuation terms quantified above are not included.

\section{Empirical Illustration}
\label{sec:empirical_filters}

This section studies the deterministic spectral estimators~\cite{logerfo2008spectral} induced by renewal laws, with the goal of isolating the predictive effect of filter shape when the estimators are implemented directly. Across both stylized experiments below, every method is applied in the eigenbasis of the training Hessian and tuned by minimizing validation-set mean squared error over the same log-spaced grid, so the comparisons are driven by filter geometry rather than by asymmetric tuning. We focus on the Gamma family from Section~\ref{sec:renewal_main}, whose filter
\[
g_{k,\tau}(\mu)
=
1-\frac{1-(1+\tau\mu/k)^{-k}}{\tau\mu}
\]
interpolates from ridge at $k=1$ to the periodic filter in the limit $k\to\infty$. All reported methods within a given experiment use identical training, validation, and test splits, and each one-parameter family is tuned on the same validation split using the same number and range of candidate hyperparameter values. In the spiked-covariance sweep, each plotted point averages 200 independent trials and tunes ridge and renewal strengths over 180 log-spaced values in \([10^{-3},10^2]\). In the block-covariance sweep in Figure~\ref{fig:block_filters}, each plotted point averages 60 independent trials and tunes each Gamma-family strength over 60 log-spaced values in the same interval; that panel is intended as a stylized renewal-only mechanism check. Appendix~\ref{app:sharp_cutoff_baseline} reports the corresponding \(B=8\) comparison that includes the sharp-cutoff baseline.

\subsection{Spiked covariance}
\label{sec:empirical_spiked}

Our first illustration uses a standard high-dimensional spiked covariance model, where a small number of informative directions sit above a broad noisy bulk in the sample spectrum \cite{donoho2018optimal}. We generate covariates from
\[
\Sigma
=
I + \sum_{i=1}^2 (\ell_i-1)u_i u_i^\top,
\qquad
(\ell_1,\ell_2)=(12,5),
\]
and place all signal in the spike directions, with population coefficients $(2,1)$ in that basis. The training size is fixed at $n_{\mathrm{train}}=80$, the validation set has size $n_{\mathrm{val}}=800$, the test set has size $5000$, and the label noise standard deviation is $\sigma_\eta=1$. We then vary the number of features \(p\), so that the aspect ratio \(\gamma=p/n_{\mathrm{train}}\) ranges from \(0.25\) to \(3.75\). Signal is confined to the two spike directions, while varying \(\gamma\) changes the size and spread of the sample-spectrum bulk formed by directions with zero true coefficient, where fitted variation is driven by noise rather than predictive signal.

\begin{figure}[htbp]
\centering
\includegraphics[width=\textwidth]{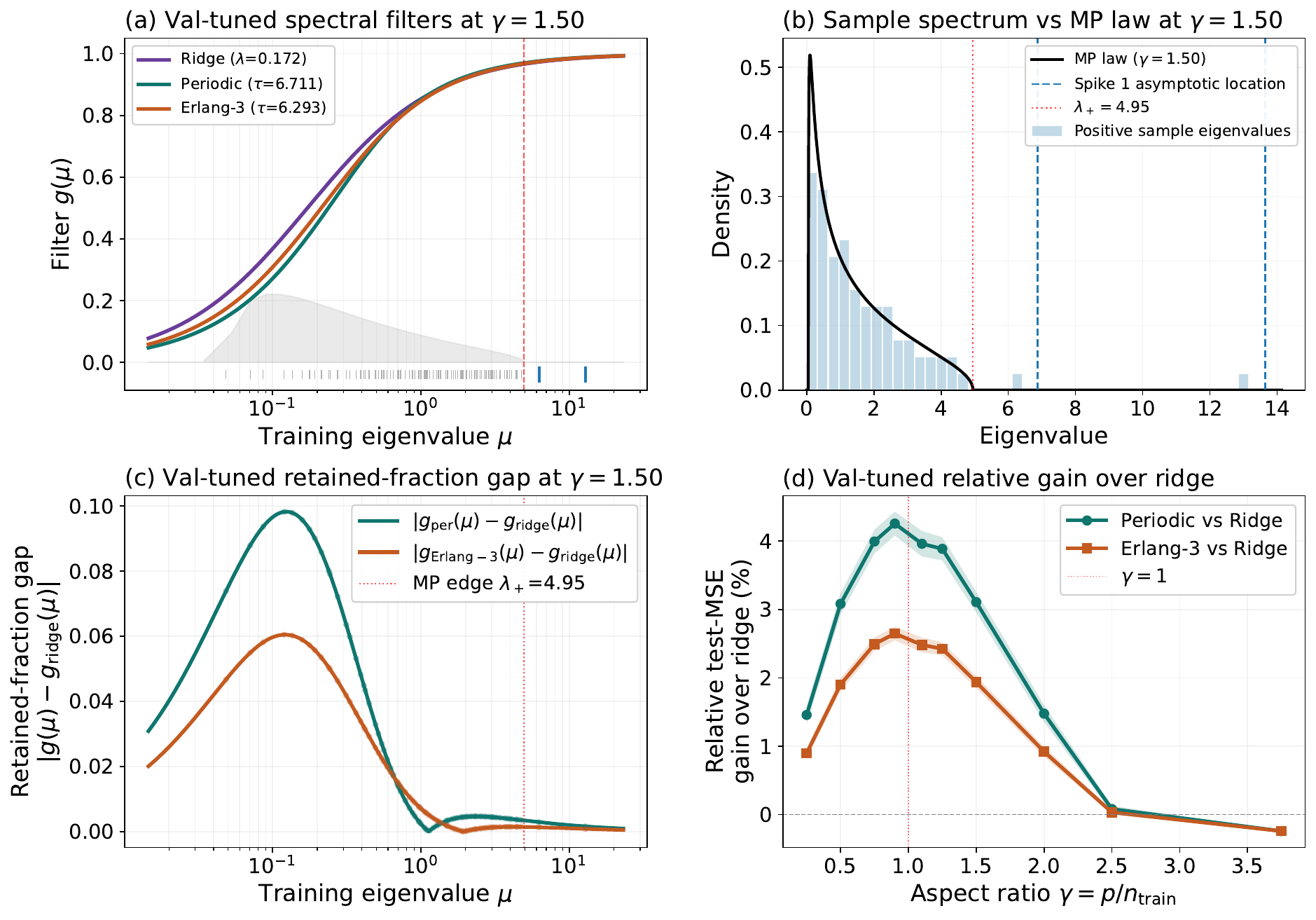}
\caption{\textbf{Validation-tuned non-exponential filters on the spiked covariance model.}
\textbf{(a)} Representative filter curves at \(\gamma=1.5\), with periodic and Erlang-3 filters each using its own validation-tuned mean reset interval. The gray fill shows the Marchenko-Pastur continuous density, blue ticks mark positive sample spikes, gray ticks mark positive sample bulk eigenvalues, and the red dashed line marks the Marchenko-Pastur edge.
\textbf{(b)} Sample spectrum compared with the Marchenko-Pastur law, including the point mass at zero when \(p>n_{\mathrm{train}}\).
\textbf{(c)} Absolute retained-fraction gap from ridge, \(|g(\mu)-g_{\mathrm{ridge}}(\mu)|\), as a function of the training eigenvalue for the same representative problem.
\textbf{(d)} Relative test-MSE gain over ridge as a function of the aspect ratio \(\gamma=p/n_{\mathrm{train}}\). Shaded bands show 95\% Monte Carlo intervals computed as \(\pm 1.96\) paired standard errors for the gain over ridge.}
\label{fig:spiked_filters}
\end{figure}

Figure~\ref{fig:spiked_filters} shows the resulting behaviour. In the representative overparameterized case \(\gamma=1.5\) (\(p=120\), \(n_{\mathrm{train}}=80\)), the sample spectrum has a visible spike-bulk structure together with a point mass at zero. Panel~(a) shows that the validation-tuned non-exponential filters are more selective than ridge on the weak and intermediate part of the spectrum. Periodic resetting gives the strongest additional shrinkage in this range, while Erlang-3 stays closer to periodic than to ridge. On the largest eigenvalues, the three filters nearly coincide, so the signal-carrying directions are largely preserved. Panel~(c) confirms this mode by mode: the retained-fraction gap from ridge is largest on weak and intermediate modes and becomes very small on the strongest modes.

Panel~(d) shows that this difference in filter shape translates into an averaged prediction effect over a range of aspect ratios. Periodic improves on ridge by roughly $1$-$4\%$ through the regime with low and intermediate \(\gamma\), while Erlang-3 yields smaller but still positive gains over much of the same range. For example, at \(\gamma=1.5\), the validation-tuned mean test MSEs are approximately \(2.322\) for ridge, \(2.250\) for periodic, and \(2.277\) for Erlang-3, with Monte Carlo standard errors for each method about \(0.023\).

\subsection{Block covariance}
\label{sec:empirical_block}

Our second illustration is based on correlated feature modules that exhibit a block-covariance structure. We partition the covariates into one predictive block of size six and $B$ nuisance blocks of size eight. Within-block correlations are $\rho_{\mathrm{sig}}=0.80$ for the signal block and $\rho_{\mathrm{nui}}=0.45$ for the nuisance blocks, with weak cross-block correlation $0.02$. Only the first block carries true signal: the coefficient vector is \((1,0.9,0.8,0.7,0.6,0.5)\) on that block and zero on every nuisance block. The training, validation, and test sizes are $(60,180,1200)$ with noise level $\sigma_\eta=2$. The nuisance blocks alter the covariance structure without contributing to the response, so increasing \(B\) adds correlated non-predictive directions while leaving the predictive block unchanged.

\begin{figure}[htbp]
\centering
\includegraphics[width=\textwidth]{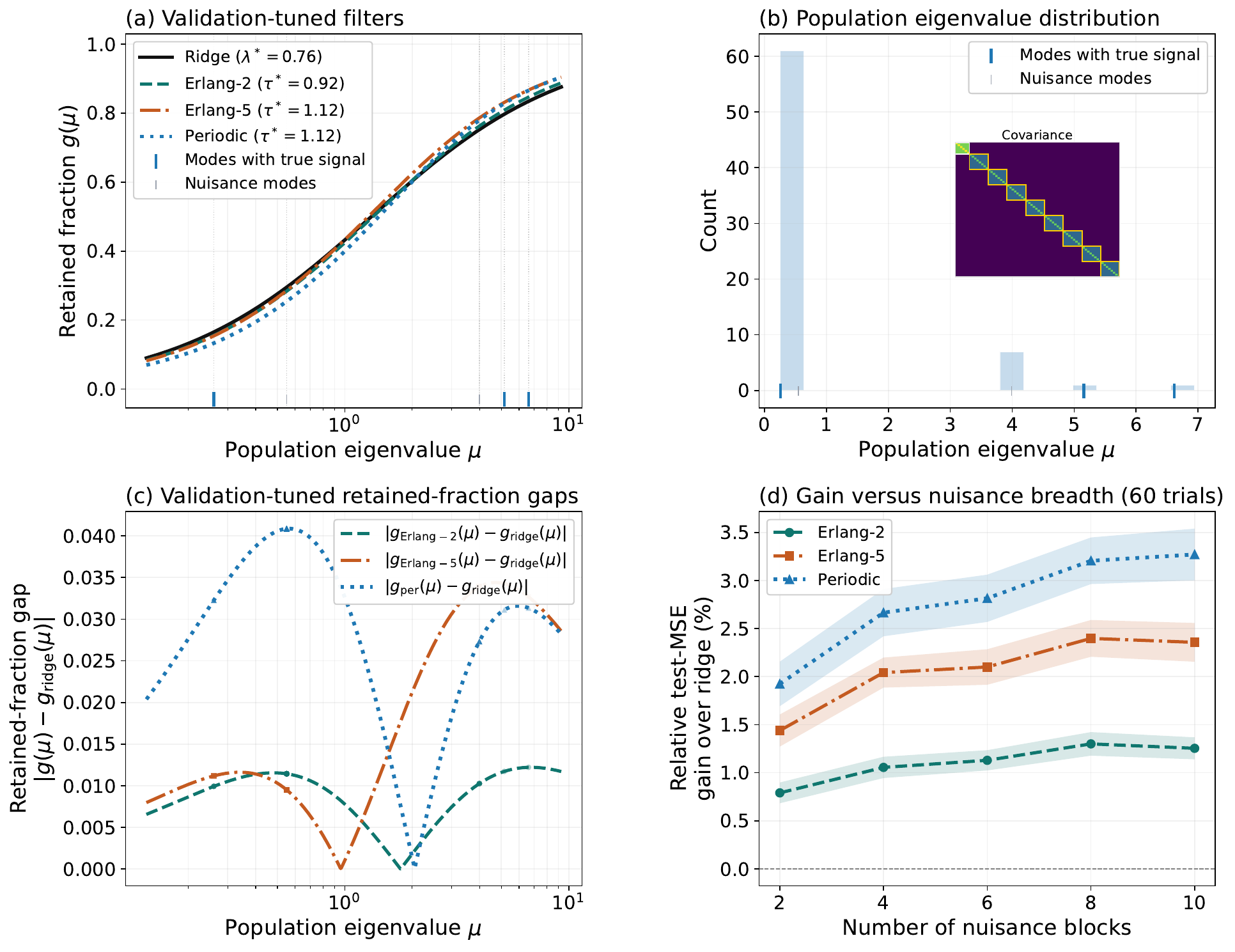}
\caption{\textbf{Block-covariance example with correlated nuisance features.}
\textbf{(a)} Validation-tuned Gamma-family filters on a representative split.
\textbf{(b)} Population eigenvalue distribution for the representative model, with signal-carrying modes separated from nuisance modes. The inset shows the underlying block covariance matrix, with the predictive block outlined in white and nuisance blocks in yellow.
\textbf{(c)} Retained-fraction gap from ridge, \(|g(\mu)-g_{\mathrm{ridge}}(\mu)|\), across the spectrum.
\textbf{(d)} Relative test-MSE gain over ridge as the number of nuisance blocks increases. Shaded bands show 95\% Monte Carlo intervals computed as \(\pm 1.96\) paired standard errors for the gain over ridge.}
\label{fig:block_filters}
\end{figure}

Figure~\ref{fig:block_filters} summarizes the resulting mechanism. Panel~(a) plots the validation-tuned Gamma-family filters on a representative split, while panel~(b) shows the corresponding population eigenvalue distribution together with the signal-carrying modes. The true signal lives on a small subset of the spectrum, while nuisance modes occupy a broader lower-eigenvalue band. On the weak part of the spectrum, the fitted curves follow the renewal interpolation qualitatively: ridge ($k=1$) is the least selective, periodic ($k=\infty$) is the most selective, and Erlang-2 and Erlang-5 lie in between. Each filter family is tuned separately on validation data. The resulting curves still follow the expected ordering on weak modes, while on the strongest modes they become extremely close and can cross slightly. Panel~(c) again shows that the largest departures from ridge occur on weak and intermediate modes.

Panel~(d) shows how that difference translates into prediction as the number of nuisance blocks increases. The gains are again modest, but consistent. They also follow the ordering by reset-time regularity: periodic performs best, Erlang-5 is intermediate, and Erlang-2 stays closest to ridge. The improvement is largest at intermediate values of \(B\) and becomes less pronounced at the largest values shown. So, the block model supports the same mechanism as the spiked-covariance example, but in a setting where the distracting directions come from correlated feature modules.

In both experiments, the non-exponential filters differ most from ridge on weak and intermediate eigenmodes and improve test MSE modestly after validation tuning. In the spiked model, these modes belong mainly to sample-spectrum directions without predictive signal, whereas in the block model they arise from correlated nuisance predictors that affect the covariance structure but not the response. Thus, in these two settings, changing the shrinkage applied to weak directions changes test risk because those directions contain little or no response signal.

More generally, the renewal-induced filters considered here act by sharpening shrinkage at the weak end of the spectrum while leaving the strongest directions almost unchanged. Relative to ridge, non-exponential laws such as Erlang and the periodic limit suppress small and intermediate eigenvalues more aggressively, but their retained fractions become very close to ridge on large eigenvalues. This makes them potentially advantageous when predictive signal is concentrated in a few leading eigendirections and a broader lower-eigenvalue band is dominated by estimation noise or nuisance structure. By contrast, when signal is spread substantially into weak directions, or when there is no clear separation between leading signal directions and the lower-eigenvalue noise-dominated directions, the same extra selectivity is more likely to be counterproductive and ridge remains preferable. In other words, the gain is controlled jointly by spectral shape, signal alignment, and noise level. Appendix~\ref{app:sharp_cutoff_baseline} places these renewal-filter gains next to a classical PCR/TSVD-style sharp cutoff (principal component regression / truncated singular value decomposition). In the reported spiked and block examples, the cutoff is stronger; so the empirical claim here is that smooth renewal-admissible filters improve over ridge and have a reset-time interpretation, while hard thresholding can be more aggressive and more effective in these settings.

\section{Conclusion and discussion}
\label{sec:conclusion_discussion}

Starting from the Laplace-average representation of ridge, this paper interprets that identity through stochastic resetting. For linear quadratic gradient flow with Poisson resetting, the exponential averaging time is the stationary age since the last reset, and the stationary mean is \((H+rI)^{-1}b\), which is the ridge estimator with penalty \(r\).

The Poisson identity is one special case of a more general relationship between reset timing and spectral shrinkage. Under Poisson resetting, the time elapsed since the most recent reset has an exponential equilibrium distribution and averaging the gradient-flow trajectory with respect to this distribution recovers the ridge filter exactly. A different renewal law produces a different age distribution and hence a different spectral filter. Such a law may agree with ridge on the finite spectrum of a particular design, but only the exponential law agrees with scalar ridge for every positive curvature. This matters when predictive signal is concentrated in a few leading eigendirections, while weaker eigendirections carry little signal and are more strongly affected by noise. In that setting, filters that shrink weak modes more strongly than ridge may improve prediction.

There is, however, an important distinction between the filter induced in the mean and the random dynamics used to realize it. At a matched mean filter, Poisson resetting has the same center as deterministic ridge, but its trajectories continue to fluctuate because of accumulated OU noise and randomness in the time since the last reset. Under the additive-noise model, these terms can only add risk, so deterministic ridge weakly dominates the dynamic implementation of Poisson resetting. Therefore, any predictive improvement obtained from non-exponential renewal filters comes from their different spectral shrinkage.

This identity at the level of the mean is closely related to the classical Bayesian interpretation of ridge regression~\cite{mackay1992bayesian,bishop2006pattern}. In fixed-design Gaussian linear regression with isotropic Gaussian prior \(w\sim\mathcal N(w_0,\tau^2 I)\) and observation noise variance \(\sigma_\eta^2\), the posterior over \(w\) is itself Gaussian, so its posterior mean and posterior mode coincide at \((X^\top X+\lambda I)^{-1}(X^\top y+\lambda w_0)\), with \(\lambda=\sigma_\eta^2/\tau^2\) up to the loss-normalization convention. Comparing with Section~\ref{sec:linear}, resetting to \(w_0\) produces exactly the same center when \(r=\lambda\). The similarity, however, is only at the level of the estimator center. Bayesian ridge obtains the shifted normal equations from Gaussian-prior conjugacy, whereas Poisson resetting obtains the same shift by averaging the gradient-flow path from the reset point over an exponential age distribution. In other words, stochastic resetting is not the Bayesian posterior itself, but a dynamical mechanism whose stationary mean reproduces the posterior center.

The exact results of the paper are obtained in a deliberately tractable setting, namely continuous-time linear quadratic dynamics with isotropic resetting and an approximation in which the stochastic noise has constant diffusion. This makes the Poisson-ridge identity and the covariance formulas explicit, but it also limits how directly the fluctuation analysis should be transferred to practical SGD. In mini-batch SGD, the noise covariance typically depends on the current parameter, the batch size, and the data geometry, so the Ornstein-Uhlenbeck approximation used here captures only part of the full stochastic structure; see, for example, Ali, Dobriban, and Tibshirani~\cite{ali2020implicit}. In future work, one could replace constant diffusion by state-dependent gradient noise. More broadly, the optimization problem over general renewal laws remains open, as do discrete-time effects, anisotropic resetting, and extensions beyond the linear regime.

\clearpage

\appendix

\section*{Appendix}

\section{Derivations for Section~\ref{sec:linear}}
\label{app:pedagogical}

This appendix section presents the step-by-step derivations that were omitted from Section~\ref{sec:linear} for brevity. Throughout this appendix, \(H=X^\top X\), \(b=X^\top y\), resets return the process to \(0\) at Poisson rate \(r>0\), and \(m(t):=\E[w(t)]\).

\subsection{Mean evolution via one-step conditioning}
\label{app:pedagogical_mean}

We derive the mean ODE \eqref{eq:mean_dynamics_ped} by conditioning on whether a reset occurs in a small interval \([t,t+\Delta t]\):
\begin{equation*}
m(t+\Delta t)
= \underbrace{(r\Delta t)\cdot 0}_{\text{reset contribution}}
+ \underbrace{(1-r\Delta t)\,\E\!\left[w(t+\Delta t)\mid \text{no reset in }[t,t+\Delta t]\right]}_{\text{no-reset contribution}}
+ O(\Delta t^2).
\end{equation*}
Under no reset, the state follows the ODE \eqref{eq:gf_linear_ped}. To first order in \(\Delta t\),
\[
w(t+\Delta t)=w(t)+(-Hw(t)+b)\Delta t + O(\Delta t^2).
\]
Because the Poisson reset event over \([t,t+\Delta t]\) is independent of the current state, conditioning on no reset does not change the distribution of \(w(t)\). Thus, using linearity of expectation,
\[
\E\!\left[w(t+\Delta t)\mid \text{no reset}\right]
= m(t)+(-H m(t)+b)\Delta t + O(\Delta t^2).
\]
Substituting:
\begin{align*}
m(t+\Delta t)
&=(1-r\Delta t)\Big(m(t)+(-Hm(t)+b)\Delta t\Big)+O(\Delta t^2) \\
&= m(t)+(-Hm(t)+b)\Delta t - r m(t)\Delta t + O(\Delta t^2),
\end{align*}
because the product \((r\Delta t)\cdot((-Hm+b)\Delta t)\) is \(O(\Delta t^2)\) and is therefore discarded in the first-order limit.

Now subtract \(m(t)\) from both sides and divide by \(\Delta t\):
\[
\frac{m(t+\Delta t)-m(t)}{\Delta t} = -H m(t)+b-rm(t)+O(\Delta t).
\]
Taking \(\Delta t\to 0\), we obtain the mean ODE:
\[
\frac{dm}{dt}=-(H+rI)m+b.
\]
Since \(H\succeq 0\) and \(r>0\), \(H+rI\) is invertible. At stationarity, \(dm/dt=0\), and hence
\[
m_\infty=(H+rI)^{-1}b=(X^\top X+rI)^{-1}X^\top y.
\]

\subsection{Ridge regression solution}
\label{app:pedagogical_ridge}

Ridge regression solves the penalized least-squares problem~\cite{james2013introduction,hastie2009elements}
\[
\min_w \left\{ \frac12\norm{Xw-y}^2 + \frac{\lambda}{2}\norm{w}^2 \right\}.
\]
The gradient of the ridge objective is
\[
X^\top(Xw-y)+\lambda w = (X^\top X + \lambda I)w - X^\top y.
\]
Setting this gradient to zero gives the normal equations
\[
(X^\top X + \lambda I)w_{\text{ridge}} = X^\top y,
\]
and therefore
\[
w_{\text{ridge}} = (X^\top X + \lambda I)^{-1}X^\top y.
\]

\subsection{Spectral shrinkage derivation}
\label{app:pedagogical_spectral}

We derive equation~\eqref{eq:shrinkage} in detail. Start from the stationary mean:
\[
m_\infty=(H+rI)^{-1}b.
\]
Let the eigendecomposition of \(H\) be \(H = V\Lambda V^\top\) with \(\Lambda=\mathrm{diag}(\mu_1,\dots,\mu_d)\) and \(V^\top V = VV^\top = I\). Then
\begin{align*}
m_\infty
&= (V\Lambda V^\top + rI)^{-1}b
= (V\Lambda V^\top + rVV^\top)^{-1}b
= \big(V(\Lambda+rI)V^\top\big)^{-1}b
= V(\Lambda+rI)^{-1}V^\top b.
\end{align*}
In spectral coordinates \(\tilde m_\infty := V^\top m_\infty\) and \(\tilde b := V^\top b\):
\[
\tilde m_\infty = (\Lambda+rI)^{-1}\tilde b
= \mathrm{diag}\!\left(\frac{1}{\mu_1+r},\dots,\frac{1}{\mu_d+r}\right)\tilde b,
\]
so for each component \(i\),
\[
\tilde m_{\infty,i} = \frac{1}{\mu_i+r}\,\tilde b_i.
\]
Now connect \(\tilde b_i\) to the minimum-norm OLS coordinates. Let \(w^\star:=H^+b\) denote the minimum-norm OLS solution. Since \(b=X^\top y\) lies in the column space of \(H=X^\top X\), we have \(HH^+b=b\), and \(Hw^\star=b\). Rotating to the eigenbasis:
\begin{align*}
V^\top H w^\star &= V^\top b \\
\Lambda (V^\top w^\star) &= V^\top b \\
\Lambda \tilde w^\star &= \tilde b,
\end{align*}
where \(\tilde w^\star := V^\top w^\star\). Componentwise, \(\mu_i\,\tilde w^\star_i = \tilde b_i\). Substituting:
\[
\tilde m_{\infty,i}
= \frac{1}{\mu_i+r}\,\mu_i\,\tilde w_i^\star
= \frac{\mu_i}{\mu_i+r}\,\tilde w_i^\star.
\]
For \(\mu_i=0\), we have \(\tilde b_i=0\), so \(\tilde m_{\infty,i}=0\) as well; the identity holds for all modes.

\paragraph{Alternative derivation via reset-age Laplace transform.}
The same shrinkage factor can be obtained from the distribution of time since the last reset. In mode \(i\), the dynamics between resets are
\[
\frac{d\tilde w_i}{dt} = -\mu_i \tilde w_i + \tilde b_i,
\qquad \tilde w_i(0)=0 \text{ right after reset.}
\]
Solving for elapsed time \(\tau\) after reset gives \(\tilde w_i(\tau) = \tilde w_i^\star(1-e^{-\mu_i\tau})\) for \(\mu_i>0\). At stationarity under Poisson resetting, the age \(T\sim\mathrm{Exp}(r)\), so
\[
\tilde m_{\infty,i}
=\E[\tilde w_i(T)]
=\tilde w_i^\star\Big(1-\E[e^{-\mu_i T}]\Big).
\]
The Laplace transform of \(T\) is \(\E[e^{-sT}] = r/(r+s)\). Setting \(s=\mu_i\):
\[
\tilde m_{\infty,i}
= \tilde w_i^\star\left(1-\frac{r}{r+\mu_i}\right)
= \tilde w_i^\star\frac{\mu_i}{\mu_i+r},
\]
recovering equation~\eqref{eq:shrinkage}. The modal filter is \(g_T(s)=1-\mathcal L_T(s)=s/(s+r)\), and evaluating at \(s=\mu_i\) yields \(\mu_i/(\mu_i+r)\).

\section{General Renewal Resetting: Mean Filters}
\label{app:renewal_mean_filters}

Poisson resetting is only one member of a larger renewal law family. Once the reset-time law is no longer memoryless, the continuous-time process need not admit a time-homogeneous stationary law, but there is still a canonical equilibrium snapshot law obtained by observing the process at a large time independent of the renewal process. The key random variable is then the age \(A\) since the most recent reset. For regularly timed resets, such as deterministic resetting every \(T\) units, the equilibrium snapshot is obtained by observing at a random point in the reset cycle, equivalently by averaging over one cycle. This appendix isolates the corresponding filter construction at the level of the mean. The covariance and risk extensions are developed separately in Appendix~\ref{app:renewal_cov_risk}.

Let \(S_1,S_2,\dots\) be i.i.d.\ reset intervals with common reset-time law \(S\), supported on \((0,\infty)\), and assume \(\E[S]<\infty\). Let \(\tau_n=\sum_{k=1}^n S_k\) be the renewal epochs.
For the formulas in this appendix written in terms of \(H^{-1}\), we assume \(H\succ 0\) for notational simplicity. In the general semidefinite case, the integral form \(\phi(a)=\int_0^a e^{-Hu}b\,du\) gives the corresponding mean formulas, equivalently replacing \(H^{-1}b\) by \(H^+b\) on the column space of \(H\). For the analysis of the mean, the only within-cycle object needed is the deterministic gradient-flow path started from zero,
\begin{equation*}
\phi(a):=(I-e^{-Ha})H^{-1}b,
\qquad a\ge 0.
\end{equation*}
If the process is observed at a large time independent of the renewal process, then the age \(A\) since the last reset converges in distribution to the equilibrium age law when the reset-time distribution is not restricted to regularly spaced times, as assumed in Section~\ref{sec:renewal_main}, with distribution function obtained from the renewal-reward calculation in Section~\ref{sec:renewal_main} and stated directly by Gallager~\cite{gallager1996discrete}:
\begin{equation}
\Pr(A \le a)=\frac{1}{\E[S]}\int_0^a \Pr(S>u)\,du.
\label{eq:equilibrium_age_cdf}
\end{equation}
Equivalently, for every \(\mu>0\),
\begin{equation}
\E[e^{-\mu A}]
=
\frac{1-\mathcal L_S(\mu)}{\mu\,\E[S]},
\qquad
\mathcal L_S(\mu):=\E[e^{-\mu S}],
\label{eq:equilibrium_age_laplace}
\end{equation}
which follows by integrating \eqref{eq:equilibrium_age_cdf} by parts. Thus, the mean filter is controlled by the equilibrium age law rather than directly by the reset-time law itself.

Conditioning on the equilibrium age gives the mean directly, since the equilibrium snapshot is the deterministic path evaluated at the random age. Therefore
\begin{equation}
m_\infty^{(S)} := \E[w^{(S)}] = \E[\phi(A)] = \bigl(I-\E[e^{-HA}]\bigr)H^{-1}b.
\label{eq:renewal_mean_matrix}
\end{equation}
In the eigenbasis \(H=V\Lambda V^\top\), with \(\Lambda=\mathrm{diag}(\mu_1,\dots,\mu_d)\) and \(\tilde w_i^\star=\tilde b_i/\mu_i\), this becomes
\begin{equation}
\tilde m_{\infty,i}^{(S)} = g_S(\mu_i)\,\tilde w_i^\star,
\label{eq:renewal_mean_eigen}
\end{equation}
where the induced spectral filter is
\begin{equation}
g_S(\mu)
:=
1-\frac{1-\mathcal L_S(\mu)}{\mu\,\E[S]}.
\label{eq:renewal_filter}
\end{equation}
At \(\mu=0\) we set \(g_S(0)=0\) by continuous extension, matching the nullspace convention used in the main text. At the level of the mean, resetting induces a modewise shrinkage profile determined by the Laplace transform of the reset-time law, rather than being characterized by a single regularisation strength. The mean estimator remains isotropic in the eigenbasis of \(H\), but the shape of the shrinkage curve depends on the renewal law.

The same formula also shows that exact ridge characterises the exponential reset time distribution. If one requires
\[
g_S(\mu)=\frac{\mu}{\mu+\lambda}
\qquad
\text{for all } \mu>0,
\]
then \eqref{eq:renewal_filter} implies
\[
\frac{1-\mathcal L_S(\mu)}{\mu\,\E[S]}=\frac{\lambda}{\mu+\lambda},
\qquad
1-\mathcal L_S(\mu)=\E[S]\lambda\,\frac{\mu}{\mu+\lambda}.
\]
Since \(S>0\) almost surely, \(\mathcal L_S(\mu)\to 0\) as \(\mu\to\infty\), so necessarily \(\E[S]\lambda=1\). Substituting back yields
\[
\mathcal L_S(\mu)=\frac{\lambda}{\mu+\lambda},
\]
which is exactly the Laplace transform of an exponential law. Thus, within the renewal class, the exponential reset-time distribution is the unique law that reproduces exact ridge shrinkage in every eigendirection. As in the main text, this statement requires matching for every \(\mu>0\); finite spectra can still have isolated non-exponential matches.

\subsection{Deterministic periodic resetting}
\label{app:renewal_mean_periodic}

The general mean formula becomes especially transparent in the deterministic periodic case. Set \(S\equiv T\). Then every reset interval has the same length \(T\), so observing the system at a large independent time under the random-phase convention is equivalent to observing a uniformly random phase within that interval. In other words, the equilibrium age is
\[
A=U\sim\mathrm{Unif}(0,T).
\]
Substituting this age law into the general renewal mean gives
\begin{align*}
\bar m_T := \E[w_{\mathrm{per}}]
&=
\E[\phi(U)]
=
\frac{1}{T}\int_0^T \phi(u)\,du \nonumber\\
&=
\left[I-\frac{1}{T}H^{-1}(I-e^{-HT})\right]H^{-1}b,
\end{align*}
so in the eigenbasis
\begin{equation*}
\tilde{\bar m}_{T,i}
=
\tilde w_i^\star \left(1-\frac{1-e^{-\mu_i T}}{\mu_i T}\right).
\end{equation*}
Thus, it is natural to define
\begin{equation}
h_T(\mu):=\frac{1-e^{-\mu T}}{\mu T},
\qquad
g_T(\mu):=1-h_T(\mu),
\label{eq:periodic_h}
\end{equation}
so that the periodic mean filter is
\begin{equation}
g_T(\mu)
=
1-\frac{1-e^{-\mu T}}{\mu T},
\qquad
\tilde{\bar m}_{T,i}=g_T(\mu_i)\tilde w_i^\star.
\label{eq:periodic_filter}
\end{equation}
A short Taylor expansion makes the mismatch with ridge explicit. For small \(\mu\),
\[
g_T(\mu)
=
1-\frac{1-e^{-\mu T}}{\mu T}
=
\frac{\mu T}{2}-\frac{\mu^2T^2}{6}+O(\mu^3),
\]
whereas a ridge filter would have
\[
\frac{\mu}{\mu+\lambda}
=
\frac{\mu}{\lambda}-\frac{\mu^2}{\lambda^2}+O(\mu^3).
\]
Matching the linear term forces \(\lambda=2/T\), but then the quadratic term disagrees, \(-T^2/4\) instead of \(-T^2/6\). So periodic resetting cannot be represented by any single ridge penalty across all modes.

\subsection{Gamma/Erlang laws at fixed mean}
\label{app:renewal_mean_gamma}

The Gamma family interpolates between exponential and periodic reset-time laws while keeping the mean reset interval fixed. Let the reset-time law be Gamma with mean \(\tau>0\) and shape \(k>0\), so that
\[
\mathcal L_S(\mu)=\left(1+\frac{\tau\mu}{k}\right)^{-k},
\qquad
\E[S]=\tau.
\]
Substituting this Laplace transform into the general renewal formula gives
\begin{equation}
h_{k,\tau}(\mu)
:=
\frac{1-\left(1+\frac{\tau\mu}{k}\right)^{-k}}{\tau\mu},
\qquad
g_{k,\tau}(\mu):=1-h_{k,\tau}(\mu),
\label{eq:gamma_age_laplace}
\end{equation}
or equivalently
\begin{equation}
g_{k,\tau}(\mu)
=
1-\frac{1-\left(1+\frac{\tau\mu}{k}\right)^{-k}}{\tau\mu}.
\label{eq:gamma_filter}
\end{equation}
The corresponding mean estimator is therefore
\begin{equation*}
\tilde m_{\infty,i}^{(k,\tau)} = g_{k,\tau}(\mu_i)\,\tilde w_i^\star.
\end{equation*}
These formulas make the two endpoints explicit. When \(k=1\),
\[
g_{1,\tau}(\mu)
=
1-\frac{1}{1+\tau\mu}
=
\frac{\tau\mu}{1+\tau\mu}
=
\frac{\mu}{\mu+1/\tau},
\]
which is exactly the ridge filter induced by Poisson resetting with \(r=1/\tau\). As \(k\to\infty\), the standard limit \((1+x/k)^{-k}\to e^{-x}\) gives
\[
h_{k,\tau}(\mu)\to \frac{1-e^{-\tau\mu}}{\tau\mu},
\qquad
g_{k,\tau}(\mu)\to 1-\frac{1-e^{-\tau\mu}}{\tau\mu},
\]
which recovers the deterministic periodic filter. The Gamma/Erlang family interpolates continuously between the unique exact-ridge law and the simplest non-exponential benchmark already at the level of the mean filter.

\section{Renewal Verification of Stationary Moments}
\label{app:renewal_verification}

We now re-derive the stationary moments using the renewal structure of the process, providing an independent verification of the covariance formula from Section~\ref{sec:covariance}.

In the closed-form expressions below, the factor \(H^{-1}\) assumes \(H\succ 0\). For general \(H\succeq 0\), one should instead use the integral representation \(\phi(\tau)=\int_0^\tau e^{-Hs}b\,ds\), which is the form already used in the main text.

\noindent At stationarity, the time since the last reset is \(\tau\sim\mathrm{Exp}(r)\). Conditioned on \(\tau\), the state is simply the OU process restarted from \(w=0\) and evolved for time \(\tau\):
\begin{equation*}
w \,|\, \tau \;=\; \underbrace{(I - e^{-H\tau})H^{-1}b}_{\phi(\tau)\,\text{(deterministic)}} \;+\; \underbrace{\int_0^\tau e^{-H(\tau - s)}\sigma\,dW(s)}_{\xi(\tau)\,\text{(stochastic, mean zero)}},
\end{equation*}
where $\phi(\tau) = \int_0^\tau e^{-Hs}\,b\,ds$ for general $H$. The stochastic part \(\xi(\tau)\) is Gaussian with mean zero and covariance
\begin{equation*}
C(\tau) = \int_0^\tau e^{-Hu}\,\Snoise\,e^{-Hu}\,du.
\end{equation*}
From this representation, the stationary mean follows by averaging \(\phi(\tau)\). Since \(\phi(\tau)\) is deterministic given \(\tau\) and \(\E[\xi(\tau)\mid\tau]=0\),
\begin{align*}
\minf &= r\int_0^\infty e^{-r\tau}\,\phi(\tau)\,d\tau = r\int_0^\infty e^{-r\tau}(I - e^{-H\tau})H^{-1}b\,d\tau \nonumber \\
&= H^{-1}b - r\,H^{-1}(H + rI)^{-1}b.
\end{align*}
Using the resolvent identity $H^{-1} - rH^{-1}(H+rI)^{-1} = (H+rI)^{-1}$ (which follows from $H^{-1}[I - r(H+rI)^{-1}] = H^{-1}H(H+rI)^{-1} = (H+rI)^{-1}$), we recover $\minf = (H + rI)^{-1}b$.

The second moment follows in the same way. Because \(\phi(\tau)\) is deterministic given \(\tau\) and \(\xi(\tau)\) is conditionally Gaussian with mean zero,
\begin{equation*}
\E[ww^\top \,|\, \tau] = \phi(\tau)\phi(\tau)^\top + C(\tau).
\end{equation*}
Averaging over the exponential age law gives
\begin{equation}
\Minf = r\int_0^\infty e^{-r\tau}\bigl[\phi(\tau)\phi(\tau)^\top + C(\tau)\bigr]\,d\tau.
\label{eq:second_moment_renewal}
\end{equation}
The SGD-noise contribution is
\begin{align*}
&r\int_0^\infty e^{-r\tau}\,C(\tau)\,d\tau = r\int_0^\infty e^{-r\tau}\int_0^\tau e^{-Hu}\,\Snoise\,e^{-Hu}\,du\,d\tau.
\end{align*}
Exchanging the order of integration ($0 \leq u \leq \tau < \infty$ becomes $0 \leq u < \infty$, $\tau \geq u$):
\begin{align*}
&= r\int_0^\infty e^{-Hu}\,\Snoise\,e^{-Hu}\left(\int_u^\infty e^{-r\tau}\,d\tau\right)du \nonumber \\
&= r\int_0^\infty e^{-Hu}\,\Snoise\,e^{-Hu}\cdot\frac{e^{-ru}}{r}\,du = \int_0^\infty e^{-ru}\,e^{-Hu}\,\Snoise\,e^{-Hu}\,du.
\end{align*}
In the eigenbasis of $H$:
\begin{equation}
(\widetilde{\Sigma}^{\mathrm{(SGD)}}_\infty)_{ij} = (\tSnoise)_{ij}\int_0^\infty e^{-(\mu_i + \mu_j + r)u}\,du = \frac{(\tSnoise)_{ij}}{\mu_i + \mu_j + r}.
\label{eq:sgd_eigenbasis_renewal}
\end{equation}
This matches the SGD component from the generator approach.

The timing randomness of the resets contributes the remaining covariance. Writing \(\Sinf=\Minf-\minf\minf^\top\) and separating the deterministic and stochastic pieces of \eqref{eq:second_moment_renewal} gives
\[
\Sinf = \underbrace{r\int_0^\infty e^{-r\tau}\,C(\tau)\,d\tau}_{\Ssgd} + \underbrace{r\int_0^\infty e^{-r\tau}\,\phi(\tau)\phi(\tau)^\top\,d\tau - \minf\minf^\top}_{\Sreset}.
\]
For the reset component in the eigenbasis, with \(\tilde{\phi}_i(\tau) = \tilde{b}_i(1 - e^{-\mu_i\tau})/\mu_i\) for \(\mu_i>0\), one finds
\begin{align*}
r\int_0^\infty e^{-r\tau}\,\tilde{\phi}_i(\tau)\,\tilde{\phi}_j(\tau)\,d\tau &= \frac{\tilde{b}_i\tilde{b}_j}{\mu_i\mu_j}\,r\int_0^\infty e^{-r\tau}(1 - e^{-\mu_i\tau})(1 - e^{-\mu_j\tau})\,d\tau.
\end{align*}
Expanding the product and evaluating each Laplace integral:
\begin{equation*}
r\int_0^\infty e^{-r\tau}(1 - e^{-\mu_i\tau})(1 - e^{-\mu_j\tau})\,d\tau = 1 - \frac{r}{r+\mu_i} - \frac{r}{r+\mu_j} + \frac{r}{r+\mu_i+\mu_j}.
\end{equation*}
It remains to show that the reset variance component
\[
(\widetilde{\Sigma}^{\mathrm{(reset)}})_{ij}
= \frac{\tilde{b}_i\tilde{b}_j}{\mu_i\mu_j}
\left[
1 - \frac{r}{r+\mu_i}
  - \frac{r}{r+\mu_j}
  + \frac{r}{r+\mu_i+\mu_j}
\right]
- (\tilde{m}_\infty)_i(\tilde{m}_\infty)_j
\]
equals
\[
\frac{r\,(\tilde{m}_\infty)_i(\tilde{m}_\infty)_j}{\mu_i+\mu_j+r},
\]
as predicted by the generator. Substituting
\((\tilde{m}_\infty)_k=\tilde{b}_k/(\mu_k+r)\), the bracketed Laplace-average term can be put over the common denominator
\((r+\mu_i)(r+\mu_j)(r+\mu_i+\mu_j)\):
\begin{align*}
&1 - \frac{r}{r+\mu_i}
  - \frac{r}{r+\mu_j}
  + \frac{r}{r+\mu_i+\mu_j} \\
&= \frac{
\mu_i(r+\mu_j)(r+\mu_i+\mu_j)
- r(r+\mu_i)(r+\mu_i+\mu_j)
+ r(r+\mu_i)(r+\mu_j)}
{(r+\mu_i)(r+\mu_j)(r+\mu_i+\mu_j)}.
\end{align*}
Expanding and collecting terms, the numerator is
\[
2r\mu_i\mu_j+\mu_i^2\mu_j+\mu_i\mu_j^2
= \mu_i\mu_j(\mu_i+\mu_j+2r).
\]
Therefore,
\begin{equation*}
1 - \frac{r}{r+\mu_i}
  - \frac{r}{r+\mu_j}
  + \frac{r}{r+\mu_i+\mu_j}
= \frac{\mu_i\mu_j(\mu_i+\mu_j+2r)}
{(r+\mu_i)(r+\mu_j)(r+\mu_i+\mu_j)}.
\end{equation*}
Substituting back gives
\begin{align}
(\widetilde{\Sigma}^{\mathrm{(reset)}})_{ij}
&= \frac{\tilde{b}_i\tilde{b}_j}{(\mu_i+r)(\mu_j+r)}
\left[
\frac{\mu_i+\mu_j+2r}{\mu_i+\mu_j+r}
- 1
\right] \nonumber \\
&= \frac{\tilde{b}_i\tilde{b}_j}{(\mu_i+r)(\mu_j+r)}
\cdot \frac{r}{\mu_i+\mu_j+r} \nonumber \\
&= \frac{r\,(\tilde{m}_\infty)_i\,(\tilde{m}_\infty)_j}
{\mu_i+\mu_j+r}.
\label{eq:reset_verified}
\end{align}

\section{Classical Ridge Risk Formula}
\label{app:classical_ridge_risk}

The coefficient-risk formula \eqref{eq:ridge_risk} is classical in the ridge-regression literature. Hoerl and Kennard~\cite{hoerl1970ridge} introduced the estimator and motivated it through mean-square-error improvement over ordinary least squares. The matrix mean-square-error analysis was then developed explicitly by Theobald~\cite{theobald1974generalizations} and extended by Farebrother~\cite{farebrother1976further}. We now show how their matrix formula reduces, in the notation of this paper, to the eigenspace sum used in Section~\ref{sec:risk}.

Consider the fixed-design Gaussian linear model
\[
y = X\beta_0 + \eta,
\qquad
\eta \sim (0,\sigma_\eta^2 I),
\qquad
H := X^\top X.
\]
For ridge penalty \(\lambda > 0\), the estimator is
\[
\hat\beta_\lambda = (H+\lambda I)^{-1}X^\top y.
\]
Theobald writes the mean-square-error matrix as
\[
M(\lambda) = D(\lambda) + b(\lambda)b(\lambda)^\top,
\]
where
\[
b(\lambda) = -\lambda(H+\lambda I)^{-1}\beta_0
\]
is the bias vector and
\[
D(\lambda) = \sigma_\eta^2 (H+\lambda I)^{-1}H(H+\lambda I)^{-1}
\]
is the dispersion matrix~\cite{theobald1974generalizations}. For squared Euclidean coefficient loss, the risk is the trace of this matrix:
\begin{equation*}
R(\lambda;\beta_0)
:=
\E_\eta \norm{\hat\beta_\lambda - \beta_0}^2
=
\Tr M(\lambda)
=
\Tr D(\lambda) + \Tr\!\bigl(b(\lambda)b(\lambda)^\top\bigr).
\end{equation*}
Using \(\Tr(uu^\top) = u^\top u\), the bias contribution becomes
\begin{equation*}
\Tr\!\bigl(b(\lambda)b(\lambda)^\top\bigr)
=
\lambda^2 \beta_0^\top(H+\lambda I)^{-2}\beta_0.
\end{equation*}
Similarly, the variance contribution is
\begin{equation*}
\Tr D(\lambda)
=
\sigma_\eta^2 \Tr\!\bigl((H+\lambda I)^{-1}H(H+\lambda I)^{-1}\bigr).
\end{equation*}
Combining these identities gives the standard matrix form
\begin{equation}
R(\lambda;\beta_0)
=
\lambda^2 \beta_0^\top(H+\lambda I)^{-2}\beta_0
+
\sigma_\eta^2 \Tr\!\bigl((H+\lambda I)^{-1}H(H+\lambda I)^{-1}\bigr).
\label{eq:classical_ridge_risk_matrix_form}
\end{equation}
To pass to spectral coordinates, diagonalise \(H\) as \(H = V\Lambda V^\top\) with \(\Lambda = \mathrm{diag}(\mu_1,\dots,\mu_d)\), and define \(\alpha := V^\top\beta_0\), so that \(\alpha_i = v_i^\top\beta_0\). Because \((H+\lambda I)^{-1} = V(\Lambda+\lambda I)^{-1}V^\top\), the bias term simplifies to
\[
\beta_0^\top(H+\lambda I)^{-2}\beta_0
=
\alpha^\top(\Lambda+\lambda I)^{-2}\alpha
=
\sum_{i=1}^d \frac{\alpha_i^2}{(\mu_i+\lambda)^2}.
\]
Likewise,
\[
(H+\lambda I)^{-1}H(H+\lambda I)^{-1}
=
V(\Lambda+\lambda I)^{-1}\Lambda(\Lambda+\lambda I)^{-1}V^\top,
\]
so trace invariance under orthogonal conjugation gives
\[
\Tr\!\bigl((H+\lambda I)^{-1}H(H+\lambda I)^{-1}\bigr)
=
\Tr\!\bigl((\Lambda+\lambda I)^{-1}\Lambda(\Lambda+\lambda I)^{-1}\bigr)
=
\sum_{i=1}^d \frac{\mu_i}{(\mu_i+\lambda)^2}.
\]
Substituting these two identities into \eqref{eq:classical_ridge_risk_matrix_form} yields
\begin{equation}
R(\lambda;\beta_0)
=
\sum_{i=1}^d
\left[
\frac{\lambda^2 (v_i^\top\beta_0)^2}{(\mu_i+\lambda)^2}
+
\frac{\sigma_\eta^2 \mu_i}{(\mu_i+\lambda)^2}
\right].
\label{eq:classical_ridge_risk_spectral}
\end{equation}
Equation~\eqref{eq:ridge_risk} in the main text is precisely \eqref{eq:classical_ridge_risk_spectral} evaluated at \(\lambda = r\), because the stationary mean of the resetting process equals the ridge estimator with penalty \(\lambda=r\).

\section{Poisson Risk Monotonicity and Optimal Resetting Rate}
\label{app:poisson_optimal}

\subsection{Monotonicity of each risk component}

\noindent The three-term decomposition makes the role of \(r\) transparent. The bias term \(\mathrm{Bias}^2(r)=\sum_i r^2(v_i^\top\beta_0)^2/(\mu_i+r)^2\) increases with \(r\); the variance from observation noise, \(\mathrm{Var}_\eta(r)=\sum_i \sigma_\eta^2\mu_i/(\mu_i+r)^2\), decreases with \(r\); and the SGD term \(\Tr(\Ssgd(r))=\sum_i (\tSnoise)_{ii}/(2\mu_i+r)\) also decreases with \(r\). The reset-variance term behaves differently, starting at \(0\), rising for small \(r\), and eventually falling back to \(0\) as \(r\to\infty\).

To see the non-monotonicity, consider the \(i\)-th summand of \eqref{eq:reset_unconditional},
\[
f_i(r) = \frac{r\,c_i}{(\mu_i+r)^2(2\mu_i+r)},
\qquad
c_i := \mu_i^2(v_i^\top\beta_0)^2 + \sigma_\eta^2\mu_i \ge 0.
\]
We have $f_i(0) = 0$, $f_i(r) > 0$ for $r > 0$ whenever $c_i>0$, and $f_i(r) \to 0$ as $r \to \infty$ (the $r$ in the numerator is dominated by $r^3$ in the denominator). Since $f_i$ starts at $0$, becomes positive, and returns to $0$, it must have an interior maximum. To find it, differentiate:
\[
f_i'(r) = \frac{c_i\bigl[(2\mu_i + r)(\mu_i+r)^2 - r\{2(\mu_i+r)(2\mu_i+r) + (\mu_i+r)^2\}\bigr]}{(\mu_i+r)^4(2\mu_i+r)^2}.
\]
The numerator simplifies to $2c_i(\mu_i+r)\bigl[\mu_i^2 - \mu_i r - r^2\bigr]$. Setting the bracket to zero gives $r^2 + \mu_i r - \mu_i^2 = 0$, with positive root
\begin{equation}
r_i^* = \frac{\mu_i(\sqrt{5} - 1)}{2} \approx 0.618\,\mu_i.
\label{eq:peak_reset_var}
\end{equation}
Thus each mode's reset variance, after averaging over observation noise, peaks when the reset and gradient timescales are comparable, \(r\approx 0.62\,\mu_i\).

\subsection{Optimal resetting rate within the Poisson family}
\label{sec:optimal}

\noindent The reset rate \(r\) controls both the ridge-type mean filter and the additional fluctuation terms. A natural choice is to tune \(r\) by minimizing the objective averaged over observation noise in \eqref{eq:total_risk}. If the minimiser is interior, \(r^*>0\), then writing \(\mathcal{R}(r;\beta_0)=\sum_i \mathcal{R}_i(r)\) mode by mode and differentiating gives the stationarity condition
\begin{equation}
\sum_{i=1}^d \left[\frac{\partial}{\partial r}\left(\frac{r^2(v_i^\top\beta_0)^2 + \sigma_\eta^2\mu_i}{(\mu_i+r)^2}\right) - \frac{(\tSnoise)_{ii}}{(2\mu_i+r)^2} + \frac{\partial}{\partial r}\left(\frac{r[\mu_i^2(v_i^\top\beta_0)^2 + \sigma_\eta^2\mu_i]}{(\mu_i+r)^2(2\mu_i+r)}\right)\right] = 0.
\label{eq:foc}
\end{equation}
Indeed,
\[
\mathcal{R}(r;\beta_0) = \sum_i \mathcal{R}_i(r),
\]
with
\[
\mathcal{R}_i(r) = \frac{r^2(v_i^\top\beta_0)^2}{(\mu_i+r)^2} + \frac{\sigma_\eta^2\mu_i}{(\mu_i+r)^2} + \frac{(\tSnoise)_{ii}}{2\mu_i+r} + \frac{r[\mu_i^2(v_i^\top\beta_0)^2 + \sigma_\eta^2\mu_i]}{(\mu_i+r)^2(2\mu_i+r)}.
\]
Differentiating each ridge risk term:
\begin{align*}
\frac{\partial}{\partial r}\left[\frac{r^2(v_i^\top\beta_0)^2}{(\mu_i+r)^2}\right] &= \frac{2r\mu_i(v_i^\top\beta_0)^2}{(\mu_i+r)^3}, \\[4pt]
\frac{\partial}{\partial r}\left[\frac{\sigma_\eta^2\mu_i}{(\mu_i+r)^2}\right] &= -\frac{2\sigma_\eta^2\mu_i}{(\mu_i+r)^3}.
\end{align*}
Differentiating the SGD noise trace:
\[
\frac{\partial}{\partial r}\left[\frac{(\tSnoise)_{ii}}{2\mu_i+r}\right] = -\frac{(\tSnoise)_{ii}}{(2\mu_i+r)^2}.
\]
The derivative of the reset term after averaging over observation noise is the same one analysed above in the monotonicity calculation, and setting the full derivative to zero gives \eqref{eq:foc}.

\subsubsection{Comparison with optimal ridge and structural limitations}

The optimal ridge parameter $\lambda^*$ minimises $\mathrm{Risk}_{\mathrm{ridge}}(\lambda; \beta_0)$ only, without the SGD or reset variance terms:
\[
\lambda^* = \arg\min_{\lambda > 0}\sum_i\left[\frac{\lambda^2(v_i^\top\beta_0)^2}{(\mu_i+\lambda)^2} + \frac{\sigma_\eta^2\mu_i}{(\mu_i+\lambda)^2}\right].
\]

\noindent This optimizer \(r^*\) generically differs from the ridge-optimal penalty \(\lambda^*\). The extra SGD-noise term \(\Tr(\Ssgd)\) is strictly decreasing in \(r\), so it pushes the preferred reset rate upward relative to ridge. The reset-stochasticity term \(\E_\eta[\Tr(\Sreset)]\), by contrast, is non-monotone and can shift the optimum either upward or downward depending on the spectral configuration. Only in degenerate cases where the process-variance terms vanish altogether do the two optimization problems collapse to the same one.

\begin{figure}[htbp]
\centering
\includegraphics[width=\textwidth]{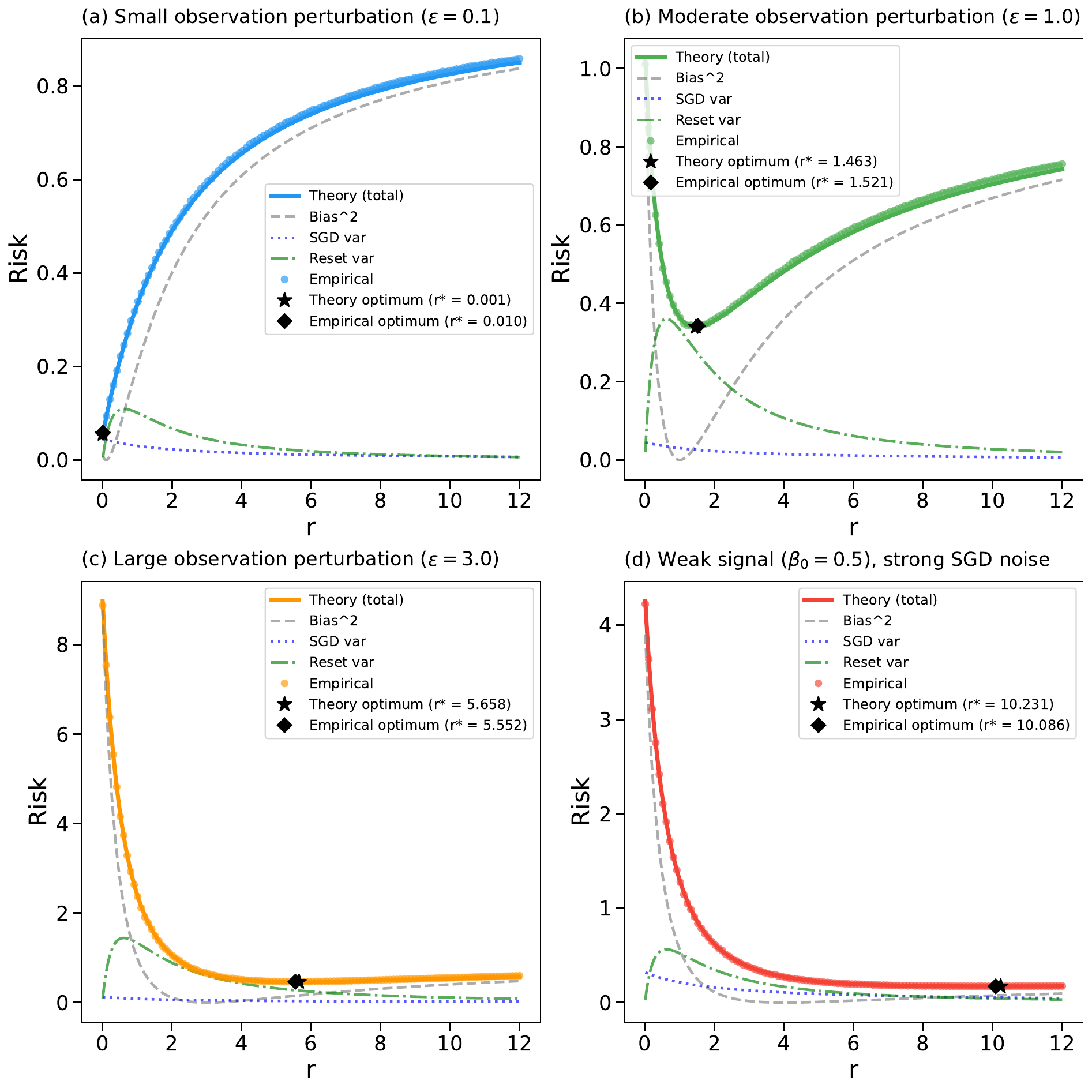}
\caption{\textbf{Scalar conditional-risk landscapes for individual observed datasets.} Each panel fixes the realised data-dependent value \(b\) and varies the reset rate \(r\). The conditional scalar risk is \(R(r) = \left(\frac{b}{\mu+r} - \beta_0\right)^2 + \frac{\sigma_n^2}{2\mu+r} + \frac{rb^2}{(\mu+r)^2(2\mu+r)}\), decomposed into bias$^2$, SGD variance, and reset variance, together with empirical estimates from stationary simulations. The black star marks the theoretical minimiser of the conditional risk and the black diamond marks the empirical minimiser. When \(b\) is only weakly perturbed from its noiseless value \(\mu\beta_0\), the optimum lies near \(r=0\); as the observation perturbation \(|b-\mu\beta_0|\) or SGD noise level increases, the optimum moves to larger \(r\). This figure illustrates dataset-to-dataset variability in the preferred reset rate; the objective defining \(r^*\) in \eqref{eq:total_risk} averages these conditional landscapes over the distribution of \(b=\mu\beta_0+\varepsilon\).}
\label{fig:optimal_reset_rate_scalar}
\end{figure}

The signal-to-noise ratio \eqref{eq:snr_ratio} typically varies strongly across modes, yet a single reset rate \(r\) acts on all of them at once. Some principal directions are rich with signal and would prefer weak regularization; others are noise-dominated and would benefit from much stronger shrinkage. Resetting inherits exactly the same isotropic limitation as ridge, which is also why the two coincide at the level of the stationary mean.

\section{General Renewal Resetting: Covariances and Risk}
\label{app:renewal_cov_risk}

The mean formulas describe how a renewal law shrinks each spectral mode. To obtain the full statistical behavior of the reset process, we must also account for variability within and across reset intervals. Keep the same reset-time law \(S\) and equilibrium age \(A\), and now suppose that between resets the state follows the additive-noise OU dynamics
\begin{equation*}
dw(t)=(-Hw(t)+b)\,dt+\sigma\,dW(t),
\qquad
t\in(\tau_n,\tau_{n+1}),
\qquad
w(\tau_n)=0,
\end{equation*}
with \(H\succ 0\) and \(\Snoise:=\sigma\sigma^\top\). If the age is fixed at \(A=a\), then the equilibrium snapshot is the deterministic mean path plus OU noise accumulated over that interval:
\begin{equation*}
w^{(S)} \,\big|\, A=a
\;=\;
\underbrace{\phi(a)}_{\text{mean path}}
\;+\;
\underbrace{\int_0^a e^{-H(a-s)}\sigma\,dW(s)}_{\xi(a)\,\text{(OU noise accumulation)}}.
\end{equation*}
The random term \(\xi(a)\) is Gaussian with mean zero, and It\^o isometry gives the conditional covariance
\begin{equation*}
C(a)=\int_0^a e^{-Hs}\,\Snoise\,e^{-Hs}\,ds.
\end{equation*}
The analogue for fluctuations then follows from the law of total covariance. Writing \(\Sigma^{(S)}:=\Cov(w^{(S)})\),
\begin{equation}
\Sigma^{(S)}
=
\underbrace{\E[C(A)]}_{\Sigma_S^{\mathrm{(SGD)}}}
\;+\;
\underbrace{\Cov(\phi(A))}_{\Sigma_S^{\mathrm{(timing)}}}.
\label{eq:renewal_cov_decomp_general}
\end{equation}
In the eigenbasis of \(H\), it is convenient to define
\begin{equation}
h_S(\mu):=\E[e^{-\mu A}]=\frac{1-\mathcal L_S(\mu)}{\mu\,\E[S]},
\qquad
g_S(\mu)=1-h_S(\mu).
\label{eq:renewal_h_general}
\end{equation}
The two covariance components then take the exact form
\begin{align}
\bigl(\widetilde{\Sigma}_S^{\mathrm{(SGD)}}\bigr)_{ij}
&=
(\tSnoise)_{ij}\,\frac{g_S(\mu_i+\mu_j)}{\mu_i+\mu_j},
\label{eq:renewal_sgd_component_general}\\[6pt]
\bigl(\widetilde{\Sigma}_S^{\mathrm{(tim)}}\bigr)_{ij}
&=
\tilde w_i^\star \tilde w_j^\star
\left[
h_S(\mu_i+\mu_j)-h_S(\mu_i)\,h_S(\mu_j)
\right].
\label{eq:renewal_timing_component_general}
\end{align}
Consequently, the exact per-mode snapshot variance is
\begin{equation}
\Var[\tilde w_i^{(S)}]
=
\frac{(\tSnoise)_{ii}\,g_S(2\mu_i)}{2\mu_i}
\;+\;
(\tilde w_i^\star)^2\Bigl[h_S(2\mu_i)-h_S(\mu_i)^2\Bigr].
\label{eq:renewal_diag_var_general}
\end{equation}
These expressions come directly from the conditional decomposition above. The term \(\E[C(A)]\) is the OU noise accumulated over a random age, while \(\Cov(\phi(A))\) is pure timing variance from observing deterministic relaxation at a random point in the reset interval. Changing the reset-time law changes the mean filter \(g_S\) and, at the same time, changes the fluctuation coefficients \(g_S(2\mu)/(2\mu)\) and \(h_S(2\mu)-h_S(\mu)^2\).

The risk after averaging over observation noise also has a closed form. If \(y=X\beta_0+\eta\) with \(\eta\sim(0,\sigma_\eta^2 I)\) and \(\alpha_i:=v_i^\top\beta_0\), then because \(\tilde w_i^\star=\alpha_i + (v_i^\top X^\top\eta)/\mu_i\), one has \(\E_\eta[\tilde w_i^\star]=\alpha_i\) and \(\Var_\eta(\tilde w_i^\star)=\sigma_\eta^2/\mu_i\). Thus, the risk of the mean estimator after averaging over observation noise is
\begin{equation}
\mathrm{Risk}_{\mathrm{mean},S}(\beta_0)
=
\sum_{i=1}^d
\left[
h_S(\mu_i)^2\,\alpha_i^2
\;+\;
\frac{\sigma_\eta^2\,g_S(\mu_i)^2}{\mu_i}
\right].
\label{eq:renewal_mean_risk_general}
\end{equation}
Adding the trace of the covariance decomposition gives the full equilibrium-snapshot risk
\begin{equation}
\mathrm{Risk}_{\mathrm{snap},S}(\beta_0)
=
\mathrm{Risk}_{\mathrm{mean},S}(\beta_0)
\;+\;
\sum_{i=1}^d \frac{(\tSnoise)_{ii}\,g_S(2\mu_i)}{2\mu_i}
\;+\;
\sum_{i=1}^d
\left(\alpha_i^2+\frac{\sigma_\eta^2}{\mu_i}\right)
\Bigl[h_S(2\mu_i)-h_S(\mu_i)^2\Bigr].
\label{eq:renewal_total_risk_general}
\end{equation}
For \(\mu_i=0\), the terms involving \(1/\mu_i\) are read by continuous extension provided \(\E[A]<\infty\); then \(g_S(2\mu_i)/(2\mu_i)\to \E[A]\), so the nullspace contribution is \(\alpha_i^2+(\tSnoise)_{ii}\E[A]\). These formulas isolate the two fluctuation terms absent from the mean-filter analysis: the coefficient for accumulated OU noise, \(g_S(2\mu)/(2\mu)\), and the coefficient for reset timing variance, \(h_S(2\mu)-h_S(\mu)^2\). The former would remain even with a fixed age; the latter is created by randomness in the reset times.

\subsection{Deterministic periodic resetting: covariance and risk}
\label{app:periodic_resetting}

Deterministic periodic resetting is the simplest non-exponential reset-time law and the cleanest benchmark for seeing how the general renewal formulas depart from ridge. Appendix~\ref{app:renewal_mean_periodic} already derived the mean filter \(g_T\) and the corresponding mean estimator. We next evaluate the general covariance and risk formulas under deterministic periodic resetting.

Under deterministic periodic resetting, \(S\equiv T\), observation at a uniformly random phase gives \(A=U\sim\mathrm{Unif}(0,T)\). The resulting random-phase state is denoted by \(w_{\mathrm{per}}\). Using the mean-filter functions \(h_T\) and \(g_T\) from \eqref{eq:periodic_h}-\eqref{eq:periodic_filter}, its covariance decomposes as
\begin{equation*}
\Sigma_T := \Cov(w_{\mathrm{per}})
=
\underbrace{\E[C(U)]}_{\Sigma_T^{\mathrm{(SGD)}}}
\;+\;
\underbrace{\Cov(\phi(U))}_{\Sigma_T^{\mathrm{(phase)}}},
\end{equation*}
with components
\begin{align*}
\bigl(\widetilde{\Sigma}_T^{\mathrm{(SGD)}}\bigr)_{ij}
&=
(\tSnoise)_{ij}\,\frac{g_T(\mu_i+\mu_j)}{\mu_i+\mu_j},
\\[6pt]
\bigl(\widetilde{\Sigma}_T^{\mathrm{(phase)}}\bigr)_{ij}
&=
\tilde w_i^\star \tilde w_j^\star
\left[
h_T(\mu_i+\mu_j)-h_T(\mu_i)\,h_T(\mu_j)
\right].
\end{align*}
Consequently,
\begin{equation*}
\Var[\tilde w_{\mathrm{per},i}]
=
\underbrace{\frac{(\tSnoise)_{ii}\,g_T(2\mu_i)}{2\mu_i}}_{\text{SGD piece}}
\;+\;
\underbrace{(\tilde w_i^\star)^2\bigl[h_T(2\mu_i)-h_T(\mu_i)^2\bigr]}_{\text{phase piece}}.
\end{equation*}
If \(y=X\beta_0+\eta\) with \(\eta\sim(0,\sigma_\eta^2 I)\), then the risk of the mean estimator is
\begin{equation*}
\mathrm{Risk}_{\mathrm{mean},T}(\beta_0)
=
\sum_{i=1}^d
\left[
h_T(\mu_i)^2\,(v_i^\top\beta_0)^2
\;+\;
\frac{\sigma_\eta^2\,g_T(\mu_i)^2}{\mu_i}
\right],
\end{equation*}
and the full random-phase risk after averaging over observation noise is
\begin{equation}
\mathrm{Risk}_{\mathrm{per}}(T;\beta_0)
=
\mathrm{Risk}_{\mathrm{mean},T}(\beta_0)
\;+\;
\Tr\!\bigl(\Sigma_T^{\mathrm{(SGD)}}\bigr)
\;+\;
\sum_{i=1}^d
\left[
\left((v_i^\top\beta_0)^2+\frac{\sigma_\eta^2}{\mu_i}\right)
\bigl(h_T(2\mu_i)-h_T(\mu_i)^2\bigr)
\right].
\label{eq:periodic_total_risk}
\end{equation}
Equivalently, if one writes the phase covariance conditionally through the phase-covariance expression above, its trace must be averaged over \(\eta\) before it is combined with \(\mathrm{Risk}_{\mathrm{mean},T}\). These formulas are exactly the general renewal expressions evaluated for \(\mathcal L_S(\mu)=e^{-\mu T}\) and \(\E[S]=T\). Equation~\eqref{eq:periodic_total_risk} is the periodic-reset analogue of the Poisson three-term decomposition: risk of the mean estimator plus SGD variance plus reset timing variance. The key difference is that the induced filter now comes from a non-exponential renewal law.

\subsection{Gamma/Erlang laws at fixed mean: covariance and risk}

The Gamma family gives an interpolation between exponential and periodic reset-time laws while keeping the mean reset interval fixed. Appendix~\ref{app:renewal_mean_gamma} already derived the corresponding mean filters \(h_{k,\tau}\) and \(g_{k,\tau}\). We now insert those same functions into the fluctuation formulas.

The exact covariance components are
\begin{align*}
\bigl(\widetilde{\Sigma}_{k,\tau}^{\mathrm{(SGD)}}\bigr)_{ij}
&=
(\tSnoise)_{ij}\,\frac{g_{k,\tau}(\mu_i+\mu_j)}{\mu_i+\mu_j},
\\[6pt]
\bigl(\widetilde{\Sigma}_{k,\tau}^{\mathrm{(tim)}}\bigr)_{ij}
&=
\tilde w_i^\star \tilde w_j^\star
\left[
h_{k,\tau}(\mu_i+\mu_j)-h_{k,\tau}(\mu_i)\,h_{k,\tau}(\mu_j)
\right].
\end{align*}
The risk of the mean estimator after averaging over observation noise is
\begin{equation*}
\mathrm{Risk}_{\mathrm{mean},k,\tau}(\beta_0)
=
\sum_{i=1}^d
\left[
h_{k,\tau}(\mu_i)^2\,(v_i^\top\beta_0)^2
\;+\;
\frac{\sigma_\eta^2\,g_{k,\tau}(\mu_i)^2}{\mu_i}
\right],
\end{equation*}
and the full snapshot risk admits the exact modewise representation
\begin{equation}
\begin{aligned}
\mathrm{Risk}_{\mathrm{snap},k,\tau}(\beta_0)
=\;&
\mathrm{Risk}_{\mathrm{mean},k,\tau}(\beta_0)
\;+\;
\sum_{i=1}^d \frac{(\tSnoise)_{ii}\,g_{k,\tau}(2\mu_i)}{2\mu_i} \\
&\;+\;
\sum_{i=1}^d
\left[
\left((v_i^\top\beta_0)^2+\frac{\sigma_\eta^2}{\mu_i}\right)
\Bigl(h_{k,\tau}(2\mu_i)-h_{k,\tau}(\mu_i)^2\Bigr)
\right].
\end{aligned}
\label{eq:gamma_snapshot_risk}
\end{equation}
These formulas are obtained simply by inserting the Gamma functions for the mean from \eqref{eq:gamma_age_laplace}-\eqref{eq:gamma_filter} into the general renewal covariance and risk expressions. Because the fluctuation formulas are built from the same \(h\)- and \(g\)-functions, they inherit the same two endpoints: the exponential law corresponding to Poisson resetting at \(k=1\) and the periodic law in the limit \(k\to\infty\). The exponential-to-periodic interpolation is explicit at the level of the mean filter, covariance, and total risk.

A non-exponential renewal law changes both the mean-filter coefficient \(g_S(\mu)\) and the fluctuation coefficients \(g_S(2\mu)/(2\mu)\) and \(h_S(2\mu)-h_S(\mu)^2\). Thus, more regular reset-time laws can help when weak modes are dominated by observation or SGD noise, because they suppress small-\(\mu\) directions more aggressively than the exponential law while leaving high-curvature modes close to their unregularised values. At the same time, the timing term shows that changing the reset-time law also changes the variance injected by uncertainty in when the trajectory is observed. Poisson resetting is thus unique for exact ridge equivalence because its reset intervals are exponential, but the formulas give no reason to expect it to be uniquely optimal for total risk; that question depends on the joint spectra of signal, observation noise, and SGD noise.

\section{External sharp-cutoff spectral baseline}
\label{app:sharp_cutoff_baseline}

The empirical comparisons in Section~\ref{sec:empirical_filters} use deterministic spectral estimators. It is also useful to include one standard non-renewal benchmark from the same spectral-estimator class~\cite{logerfo2008spectral}. We use a PCR/TSVD-style sharp cutoff~\cite{massy1965principal,hansen1990truncated}
\[
g_{\mathrm{cut}}(\mu)=\mathbf 1\{\mu\ge c\},
\]
with \(c\) tuned on the same validation split and over the same number and range of candidate values as the other methods. This cutoff is deliberately external to the resetting construction: its discontinuity and lack of an equilibrium-age Laplace representation place it outside the completely monotone renewal class. Its role is to calibrate whether the observed gains over ridge reflect renewal-specific filter geometry or the broader benefit of suppressing weak spectral modes.

\begin{table}[htbp]
\centering
\small
\begin{tabular}{llccc}
\toprule
Experiment & Method & Test MSE & Gain over ridge & MC s.e. of gain \\
\midrule
Spiked, \(\gamma=1.5\) & Ridge & 2.322 & 0.000\% & -- \\
Spiked, \(\gamma=1.5\) & Periodic & 2.250 & 3.108\% & 0.077\% \\
Spiked, \(\gamma=1.5\) & Erlang-3 & 2.277 & 1.938\% & 0.045\% \\
Spiked, \(\gamma=1.5\) & Cutoff & 2.129 & 8.046\% & 0.406\% \\
\midrule
Block, \(B=8\) & Ridge & 5.584 & 0.000\% & -- \\
Block, \(B=8\) & Periodic & 5.404 & 3.206\% & 0.124\% \\
Block, \(B=8\) & Erlang-5 & 5.449 & 2.398\% & 0.098\% \\
Block, \(B=8\) & Erlang-2 & 5.510 & 1.300\% & 0.063\% \\
Block, \(B=8\) & Cutoff & 4.958 & 10.999\% & 0.667\% \\
\bottomrule
\end{tabular}
\caption{\textbf{Validation-tuned sharp-cutoff baseline.} The spiked-covariance row reports the representative overparameterized setting \(\gamma=1.5\) from the 200-trial sweep. The block-covariance row reports \(B=8\) nuisance blocks from a 60-trial sweep. The cutoff baseline provides an external sharp spectral comparator.}
\label{tab:sharp_cutoff_baseline}
\end{table}

The gain percentages and Monte Carlo standard errors in Table~\ref{tab:sharp_cutoff_baseline} are computed by first forming the relative improvement over ridge separately for each simulation run and then averaging those run-level improvements. Therefore, they can differ from the relative difference of the rounded mean MSE values shown in the table.

In the reported spiked and block settings, the renewal filters outperform ridge but not the sharp cutoff. These experiments establish a benefit over ridge for the smooth renewal filters considered here, rather than superiority over standard spectral truncation.

\bibliographystyle{unsrt}
\bibliography{bibliography}

@article{evans2011diffusion,
  title={Diffusion with stochastic resetting},
  author={Evans, Martin R and Majumdar, Satya N},
  journal={Physical review letters},
  volume={106},
  number={16},
  pages={160601},
  year={2011},
  publisher={APS},
  doi={10.1103/PhysRevLett.106.160601}
}

@article{evans2020stochastic,
  title={Stochastic resetting and applications},
  author={Evans, Martin R and Majumdar, Satya N and Schehr, Gr{\'e}gory},
  journal={Journal of Physics A: Mathematical and Theoretical},
  volume={53},
  number={19},
  pages={193001},
  year={2020},
  publisher={IOP Publishing},
  doi={10.1088/1751-8121/ab7cfe}
}

@article{mandt2017stochastic,
  title={Stochastic gradient descent as approximate bayesian inference},
  author={Mandt, Stephan and Hoffman, Matthew D and Blei, David M},
  journal={Journal of Machine Learning Research},
  volume={18},
  number={134},
  pages={1--35},
  year={2017}
}

@article{hoerl1970ridge,
  title={Ridge regression: Biased estimation for nonorthogonal problems},
  author={Hoerl, Arthur E and Kennard, Robert W},
  journal={Technometrics},
  volume={12},
  number={1},
  pages={55--67},
  year={1970},
  publisher={Taylor \& Francis},
  doi={10.1080/00401706.1970.10488634}
}

@article{theobald1974generalizations,
  title={Generalizations of mean square error applied to ridge regression},
  author={Theobald, Chris M},
  journal={Journal of the Royal Statistical Society Series B: Statistical Methodology},
  volume={36},
  number={1},
  pages={103--106},
  year={1974},
  publisher={Oxford University Press},
  doi={10.1111/j.2517-6161.1974.tb00990.x}
}

@article{farebrother1976further,
  title={Further Results on the Mean Square Error of Ridge Regression},
  author={Farebrother, RW},
  journal={Journal of the Royal Statistical Society Series B: Statistical Methodology},
  volume={38},
  number={3},
  pages={248--250},
  year={1976},
  publisher={Oxford University Press},
  doi={10.1111/j.2517-6161.1976.tb01588.x}
}

@inproceedings{ali2019continuous,
  title={A continuous-time view of early stopping for least squares regression},
  author={Ali, Alnur and Kolter, J Zico and Tibshirani, Ryan J},
  booktitle={The 22nd international conference on artificial intelligence and statistics},
  pages={1370--1378},
  year={2019},
  organization={PMLR}
}

@incollection{tibshirani2022laplace,
  title={Gradient flow, Laplace transforms, and infinitesimal steepest descent: Partial results and open directions},
  author={Tibshirani, Ryan J},
  booktitle={Mathematical Foundations of Robust and Generalizable Learning},
  series={Oberwolfach Reports},
  volume={19},
  pages={2683--2684},
  year={2022},
  publisher={EMS Press},
  doi={10.4171/OWR/2022/46},
  url={https://ems.press/content/serial-article-files/46984},
  note={Report No. 46/2022}
}

@inproceedings{ali2020implicit,
  title={The implicit regularization of stochastic gradient flow for least squares},
  author={Ali, Alnur and Dobriban, Edgar and Tibshirani, Ryan},
  booktitle={International conference on machine learning},
  pages={233--244},
  year={2020},
  organization={PMLR}
}

@techreport{rosasco2005spectral,
  title={Spectral methods for regularization in learning theory},
  author={Rosasco, Lorenzo and De Vito, Ernesto and Verri, Alessandro},
  institution={DISI, Universit{\`a} degli Studi di Genova},
  number={DISI-TR-05-18},
  year={2005}
}

@article{bauer2007regularization,
  title={On regularization algorithms in learning theory},
  author={Bauer, Frank and Pereverzev, Sergei and Rosasco, Lorenzo},
  journal={Journal of complexity},
  volume={23},
  number={1},
  pages={52--72},
  year={2007},
  publisher={Elsevier},
  doi={10.1016/j.jco.2006.07.001}
}

@article{logerfo2008spectral,
  title={Spectral algorithms for supervised learning},
  author={{Lo Gerfo}, L. and Rosasco, Lorenzo and Odone, Francesca and {De Vito}, Ernesto and Verri, Alessandro},
  journal={Neural Computation},
  volume={20},
  number={7},
  pages={1873--1897},
  year={2008},
  publisher={MIT Press},
  doi={10.1162/neco.2008.05-07-517}
}

@article{yao2007early,
  title={On early stopping in gradient descent learning},
  author={Yao, Yuan and Rosasco, Lorenzo and Caponnetto, Andrea},
  journal={Constructive approximation},
  volume={26},
  number={2},
  pages={289--315},
  year={2007},
  publisher={Springer},
  doi={10.1007/s00365-006-0663-2}
}

@article{blanchard2018optimal,
  title={Optimal adaptation for early stopping in statistical inverse problems},
  author={Blanchard, Gilles and Hoffmann, Marc and Rei{\ss}, Markus},
  journal={SIAM/ASA Journal on Uncertainty Quantification},
  volume={6},
  number={3},
  pages={1043--1075},
  year={2018},
  publisher={SIAM},
  doi={10.1137/17M1154096}
}

@article{sonthalia2024early,
  title={On regularization via early stopping for least squares regression},
  author={Sonthalia, Rishi and Lok, Jackie and Rebrova, Elizaveta},
  journal={arXiv preprint arXiv:2406.04425},
  year={2024},
  eprint={2406.04425},
  archivePrefix={arXiv},
  doi={10.48550/arXiv.2406.04425}
}

@article{dicker2017kernel,
  title={Kernel ridge vs. principal component regression: Minimax bounds and the qualification of regularization operators},
  author={Dicker, Lee H and Foster, Dean P and Hsu, Daniel},
  journal={Electronic Journal of Statistics},
  volume={11},
  number={1},
  pages={1022--1047},
  year={2017},
  doi={10.1214/17-EJS1258}
}

@article{donoho2018optimal,
  title={Optimal shrinkage of eigenvalues in the spiked covariance model},
  author={Donoho, David L and Gavish, Matan and Johnstone, Iain M},
  journal={Annals of statistics},
  volume={46},
  number={4},
  pages={1742--1778},
  year={2018},
  doi={10.1214/17-AOS1601}
}

@article{su2016differential,
  title={A differential equation for modeling Nesterov's accelerated gradient method: Theory and insights},
  author={Su, Weijie and Boyd, Stephen and Cand{\`e}s, Emmanuel J},
  journal={Journal of Machine Learning Research},
  volume={17},
  number={153},
  pages={1--43},
  year={2016}
}

@article{odonoghue2015adaptive,
  title={Adaptive restart for accelerated gradient schemes},
  author={O'Donoghue, Brendan and Cand{\`e}s, Emmanuel},
  journal={Foundations of computational mathematics},
  volume={15},
  number={3},
  pages={715--732},
  year={2015},
  publisher={Springer},
  doi={10.1007/s10208-013-9150-3}
}

@inproceedings{giselsson2014monotonicity,
  title={Monotonicity and restart in fast gradient methods},
  author={Giselsson, Pontus and Boyd, Stephen},
  booktitle={53rd IEEE Conference on Decision and Control},
  pages={5058--5063},
  year={2014},
  organization={IEEE},
  doi={10.1109/CDC.2014.7040179}
}

@article{pal2017first,
  title={First passage under restart},
  author={Pal, Arnab and Reuveni, Shlomi},
  journal={Physical review letters},
  volume={118},
  number={3},
  pages={030603},
  year={2017},
  publisher={APS},
  doi={10.1103/PhysRevLett.118.030603}
}

@article{chechkin2018random,
  title={Random search with resetting: a unified renewal approach},
  author={Chechkin, Aleksei and Sokolov, Igor M},
  journal={Physical review letters},
  volume={121},
  number={5},
  pages={050601},
  year={2018},
  publisher={APS},
  doi={10.1103/PhysRevLett.121.050601}
}

@article{richards2021comparing,
  title={Comparing classes of estimators: When does gradient descent beat ridge regression in linear models?},
  author={Richards, Dominic and Dobriban, Edgar and Rebeschini, Patrick},
  journal={arXiv preprint arXiv:2108.11872},
  year={2021},
  eprint={2108.11872},
  archivePrefix={arXiv},
  doi={10.48550/arXiv.2108.11872}
}

@article{lu2024saturation,
  title={On the saturation effects of spectral algorithms in large dimensions},
  author={Lu, Weihao and Zhang, Haobo and Li, Yicheng and Lin, Qian},
  journal={Advances in Neural Information Processing Systems},
  volume={37},
  pages={7011--7059},
  year={2024},
  doi={10.52202/079017-0225}
}

@article{starkov2023universal,
  title={Universal performance bounds of restart},
  author={Starkov, Dmitry and Belan, Sergey},
  journal={Physical Review E},
  volume={107},
  number={6},
  pages={L062101},
  year={2023},
  publisher={APS},
  doi={10.1103/PhysRevE.107.L062101}
}

@article{nikitin2024constructing,
  title={Constructing efficient strategies for the process optimization by restart},
  author={Nikitin, Ilia and Belan, Sergey},
  journal={Physical Review E},
  volume={109},
  number={5},
  pages={054117},
  year={2024},
  publisher={APS},
  doi={10.1103/PhysRevE.109.054117}
}

@inproceedings{loshchilov2017sgdr,
  title={SGDR: Stochastic Gradient Descent with Warm Restarts},
  author={Loshchilov, Ilya and Hutter, Frank},
  booktitle={International Conference on Learning Representations},
  year={2017}
}

@article{bae2025stochastic,
  title={Stochastic resetting mitigates latent gradient bias of SGD from label noise},
  author={Bae, Youngkyoung and Song, Yeongwoo and Jeong, Hawoong},
  journal={Machine Learning: Science and Technology},
  volume={6},
  number={1},
  pages={015062},
  year={2025},
  publisher={IOP Publishing},
  doi={10.1088/2632-2153/adbc46}
}

@article{meir2025first,
  author  = {Meir, Sagi and Keidar, Tommer D. and Reuveni, Shlomi and Hirshberg, Barak},
  title   = {First-Passage Approach to Optimizing Perturbations for Improved Training of Machine Learning Models},
  journal = {Machine Learning: Science and Technology},
  volume  = {6},
  number  = {2},
  pages   = {025053},
  year    = {2025},
  doi     = {10.1088/2632-2153/add8df}
}

@article{keidar2025adaptive,
  author  = {Keidar, Tommer D. and Blumer, Ofir and Hirshberg, Barak and Reuveni, Shlomi},
  title   = {Adaptive resetting for informed search strategies and the design of non-equilibrium steady-states},
  journal = {Nature Communications},
  volume  = {16},
  pages   = {7259},
  year    = {2025},
  doi     = {10.1038/s41467-025-62398-2}
}

@article{galashov2024non,
  title={Non-stationary learning of neural networks with automatic soft parameter reset},
  author={Galashov, Alexandre and Titsias, Michalis and Gy{\"o}rgy, Andr{\'a}s and Lyle, Clare and Pascanu, Razvan and Teh, Yee Whye and Sahani, Maneesh},
  journal={Advances in Neural Information Processing Systems},
  volume={37},
  pages={83197--83234},
  year={2024},
  doi={10.52202/079017-2647}
}

@inproceedings{nikishin2022primacy,
  title={The primacy bias in deep reinforcement learning},
  author={Nikishin, Evgenii and Schwarzer, Max and D'Oro, Pierluca and Bacon, Pierre-Luc and Courville, Aaron},
  booktitle={International conference on machine learning},
  pages={16828--16847},
  year={2022},
  organization={PMLR}
}

@inproceedings{barrett2020implicit,
  title={Implicit gradient regularization},
  author={Barrett, David GT and Dherin, Benoit},
  booktitle={International Conference on Learning Representations},
  year={2021},
  eprint={2009.11162},
  archivePrefix={arXiv},
  doi={10.48550/arXiv.2009.11162},
  url={https://openreview.net/forum?id=3q5IqUrkcF}
}

@article{paquette2022implicit,
  title={Implicit regularization or implicit conditioning? exact risk trajectories of sgd in high dimensions},
  author={Paquette, Courtney and Paquette, Elliot and Adlam, Ben and Pennington, Jeffrey},
  journal={Advances in Neural Information Processing Systems},
  volume={35},
  pages={35984--35999},
  year={2022}
}

@inproceedings{zou2021benefits,
  author    = {Zou, Difan and Wu, Jingfeng and Braverman, Vladimir and Gu, Quanquan and Foster, Dean P. and Kakade, Sham M.},
  title     = {The Benefits of Implicit Regularization from SGD in Least Squares Problems},
  booktitle = {Advances in Neural Information Processing Systems},
  volume    = {34},
  pages     = {5456--5468},
  year      = {2021}
}

@inproceedings{blanc2020implicit,
  author    = {Blanc, Guy and Gupta, Neha and Valiant, Gregory and Valiant, Paul},
  title     = {Implicit regularization for deep neural networks driven by an Ornstein-Uhlenbeck like process},
  booktitle = {Proceedings of Thirty Third Conference on Learning Theory},
  series    = {Proceedings of Machine Learning Research},
  volume    = {125},
  pages     = {483--513},
  year      = {2020},
  publisher = {PMLR}
}

@inproceedings{krogh1991simple,
  author    = {Krogh, Anders and Hertz, John A.},
  title     = {A Simple Weight Decay Can Improve Generalization},
  booktitle = {Advances in Neural Information Processing Systems},
  volume    = {4},
  pages     = {950--957},
  year      = {1991}
}

@book{ross2014introduction,
  author    = {Ross, Sheldon M.},
  title     = {Introduction to Probability Models},
  edition   = {11th},
  year      = {2014},
  publisher = {Academic Press},
  isbn      = {978-0-12-407948-9}
}

@book{schilling2012bernstein,
  author = {Schilling, Ren{\'e} L. and Song, Renming and Vondra{\v{c}}ek, Zoran},
  title = {Bernstein Functions: Theory and Applications},
  publisher = {De Gruyter},
  edition = {2nd},
  year = {2012},
  doi = {10.1515/9783110269338}
}

@article{mackay1992bayesian,
  title={Bayesian interpolation},
  author={MacKay, David JC},
  journal={Neural computation},
  volume={4},
  number={3},
  pages={415--447},
  year={1992},
  publisher={MIT Press},
  doi={10.1162/neco.1992.4.3.415}
}

@book{bishop2006pattern,
  title={Pattern recognition and machine learning},
  author={Bishop, Christopher M},
  series={Information Science and Statistics},
  year={2006},
  publisher={Springer},
  isbn={978-0-387-31073-2}
}

@article{hansen1990truncated,
  title={Truncated singular value decomposition solutions to discrete ill-posed problems with ill-determined numerical rank},
  author={Hansen, Per Christian},
  journal={SIAM Journal on Scientific and Statistical Computing},
  volume={11},
  number={3},
  pages={503--518},
  year={1990},
  publisher={SIAM},
  doi={10.1137/0911028}
}

@article{massy1965principal,
  title={Principal components regression in exploratory statistical research},
  author={Massy, William F},
  journal={Journal of the American Statistical Association},
  volume={60},
  number={309},
  pages={234--256},
  year={1965},
  publisher={Taylor \& Francis},
  doi={10.1080/01621459.1965.10480787}
}

@book{gallager1996discrete,
  title     = {Discrete Stochastic Processes},
  author    = {Gallager, Robert G.},
  series    = {The Springer International Series in Engineering and Computer Science},
  volume    = {321},
  publisher = {Springer},
  address   = {New York, NY},
  year      = {1996},
  doi       = {10.1007/978-1-4615-2329-1}
}

@book{james2013introduction,
  title     = {An Introduction to Statistical Learning: with Applications in R},
  author    = {James, Gareth and Witten, Daniela and Hastie, Trevor and Tibshirani, Robert},
  series    = {Springer Texts in Statistics},
  volume    = {103},
  year      = {2013},
  publisher = {Springer},
  address   = {New York, NY},
  doi       = {10.1007/978-1-4614-7138-7}
}

@book{hastie2009elements,
  title={The Elements of Statistical Learning: Data Mining, Inference, and Prediction},
  author={Hastie, Trevor and Tibshirani, Robert and Friedman, Jerome},
  year={2009},
  publisher={Springer}
}

\end{document}